\pdfoutput=1

\documentclass[11pt]{article}

\usepackage{EMNLP2023}

\usepackage{graphicx}
\usepackage{times}
\usepackage{latexsym}
\usepackage{gb4e}
\usepackage{amsmath}
\usepackage{amssymb}
\usepackage{bm}
\usepackage{paralist}
\usepackage{placeins}
\noautomath

\usepackage[T1]{fontenc}

\usepackage[utf8]{inputenc}
\usepackage[english]{babel}
\usepackage[english=british]{csquotes}

\usepackage{microtype}

\usepackage{inconsolata}

\usepackage{enumitem}
\usepackage{booktabs}
\usepackage{multirow}
\usepackage{xspace}
\usepackage[normalem]{ulem}
\useunder{\uline}{\ul}{}

\usepackage{annotates}

\title{
Investigating semantic subspaces of Transformer sentence embeddings\\
through linear structural probing
}

\author{Dmitry Nikolaev \quad Sebastian Pad{\'o} \\
        Institute for Natural Language Processing, University of Stuttgart \\
        \texttt{dnikolaev@fastmail.com} \quad  \texttt{pado@ims.uni-stuttgart.de}}

\begin{document}

\maketitle

\begin{abstract}
    The question of what kinds of linguistic information are encoded in different layers
    of Transformer-based language models is of considerable interest for the NLP community. 
    Existing work, however, has overwhelmingly focused on word-level representations and 
    encoder-only language models with the masked-token training objective. 
    In this paper, we present experiments with \textit{semantic structural probing}, 
    a method for studying sentence-level representations 
    via finding a subspace of the embedding space that provides
    suitable task-specific pairwise distances between data-points. 
    We apply our method to language models from different families (encoder-only, decoder-only, 
    encoder-decoder) and of different sizes in the context of two tasks, semantic textual similarity 
    and natural-language inference. We find that model families differ substantially in their 
    performance and layer dynamics, but that the results are largely model-size invariant.
\end{abstract}

\section{Introduction}
\label{sec:introduction}

It is more or less generally assumed that the success of bidirectional masked language models (MLMs), such as 
BERT \citep{devlin-etal-2019-bert}, on downstream tasks is largely due to the fact that in pre-training they 
learn to compute rich and well-structured representations of their inputs. 
More precisely, it is often argued that the task of masked language modelling encourages models to 
successively aggregate lexical/collocational, syntactic, and semantic information from the input text  
as the activations progress through encoder layers \citep{tenney-etal-2019-bert}. The extent to which BERT-like 
models follow the stages of the classical NLP pipeline \cite{niu-etal-2022-bert} or 
accumulate contextual information \citep{kunz-kuhlmann-2022-linguistic} has been questioned. 
However, the association of middle layers of MLMs with syntax and higher levels with semantic information is not 
widely disputed as a general principle and is taken for granted in many papers on model analysis and 
downstream applications \citep{chi-etal-2020-finding,li-etal-2021-bert,sharma2022exploratory}.

Despite the high volume of literature on Transformer representations, these studies are mostly constrained
in two ways: (i)~they deal almost exclusively with word/token-level and not sentence-level embeddings,
and (ii)~the lion's share of attention is given to encoder-only MLMs, notably variants of BERT 
\citep[cf.][]{reif2019visualizing,hewitt-manning-2019-structural,vulic-etal-2020-probing,conia-navigli-2022-probing}.
As a result, the representations computed by text-to-text models, such as T5, and causal language
models, such as members of the GPT family, remain understudied. This can be largely attributed to the 
fact that the standard way of analyzing pretrained language models, namely \textit{probing}, proceeds by 
applying  linear classifiers to token representations at different layers \citep{belinkov-2022-probing}. This 
approach is not as easily applicable to MLM-derived sentence representations or to representations 
computed by other types of models.\footnote{Cf., however, \citet{liu-etal-2019-linguistic} and works targeting 
representations computed by LSTM-based LMs: 
\citet{giulianelli-etal-2018-hood,aina-etal-2019-putting,sorodoc-etal-2020-probing,sukumaran2022lstms}, 
and others. In this work, we focus on the processing of natural language, 
cf.\ \citet{bhattamishra-etal-2020-ability} and \citet{traylor-etal-2021-mean} on the ability 
of LMs to tackle formal languages.}
Conversely, while it is possible to provide nuanced analyses 
of causal Transformer-based models \citep{geva-etal-2021-transformer,geva-etal-2022-transformer}, 
such analyses are not easily transferable to MLMs \citep{nikolaev2023universe}. 

In this study, we propose
a unified methodology for studying layer-specific sentence-level representations extracted from 
masked, 
text-to-text, and causal language models. We 
analyze these representations via \textit{structural semantic probing}, largely inspired by 
\citet{chi-etal-2020-finding}. Instead of directly predicting features of interest from 
representations, structural probing projects them onto lower-dimensional subspaces where
distances are interpretable in terms of task properties, or where different classes
of data points are directly linearly separable. By varying the dimensionality of the projection space, 
we can gauge the amount of information contained in the embeddings.

While \citet{chi-etal-2020-finding} identify well-structured syntactic subspaces, 
i.e.\ those encoding the topology and labels of Universal Dependency trees, we target sentence-level 
semantic subspaces and carry out experiments on two semantic tasks, viz.\ sentence similarity 
and natural language inference (NLI). Our contributions are as follows:

\begin{enumerate}
    \item We analyse the efficiency of solving different semantics-level downstream tasks using 
    only suitably projected sentence embeddings derived from vanilla pre-trained encoder-only, 
    encoder-decoder, and decoder-only models.
    \item We conduct an extensive analysis of the informativeness of embeddings derived from different 
    model layers using varying dimensionalities of projection subspaces. 
    Many of the models we study have never been analysed in this way, and we find that
    their behaviour is influenced in interesting ways by both architecture and training regime.
    \item We conduct our experiments at widely differing model scales: from BERT base, T5 mini, and OPT 125m to 
    T5 XXL, Llama 13B, and OPT 66b. Our main finding is that the way information is structured across layers
    is largely \textit{scale invariant}, with models sharing the same architecture and 
    training regime demonstrating similar activation patterns.
    \item We show that three major NLI datasets -- SNLI, MNLI, and ANLI -- lead to very different results
    when tackled with projected vanilla embeddings. While SNLI and MNLI, surprisingly, can be almost
    \enquote{solved} with most vanilla models, ANLI, in contrast, is nearly completely opaque, 
    and only embeddings from the biggest models are useful there.
\end{enumerate}
The structure of the paper is as follows: \S\ 2 introduces structural probing and its application; \S~3 lays out our experimental setup; \S~4 presents and discusses our findings, and \S~5 
concludes.

\section{Semantic structural probing}
\label{sec:methods}



In all our experiments, we assume that we are given a set of sentences \(s_i \in \mathcal{S}\) 
and a corresponding set of sentence representations \(r_{s_i,m,l} \in \mathcal{R}\),
where each element is indexed with a sentence, a model from which it was derived, and the model 
layer. (Model and layer subscripts will be omitted when not needed.) Depending on the 
task, we also have labels of different types either for sentence pairs 
(\(l_{i,j}\)) or individual sentences (\(l_i\)). 
We target the following tasks:

\begin{enumerate}
    \item Semantic textual similarity (STS): a pair of sentences is labelled with a number from 0 to 
    5, where 0 corresponds to the smallest degree of semantic similarity and 5 to the maximal degree. 
    We map these labels to the range [0, 1] of semantic differences.
    \item Textual entailment (TE): an ordered pair of sentences is labelled according to whether the 
    second sentence is entailed by the first one or contradicts it. To simplify
    the analysis, we do not address neutral sentence pairs.
\end{enumerate}
To study the semantic organisation of sentence representations, we aim to find a projection 
matrix  \(M\) to a lower-dimensional space, such that we can directly \enquote{read off} the answer 
from the application of the matrix to elements of \(\mathcal{R}\).


For the \textbf{STS} task, we choose an \(M\) that minimises the differences between the gold-label 
similarities and the Euclidean distance between embeddings (averaged over the mini-batch):
\begin{equation}
    \mathcal{L}_{\text{STS}} = ( || Mr_{s_i} - Mr_{s_j} ||_2 - l_{i,j} )^2    
\end{equation}
This corresponds to learning an approximation to the Mahalanobis
matrix \(M^{T}M\), that is, to learning a distance metric in the embedding
space. (This interpretation carries over to our other experiments.)
This distance metric is optimised to correlate well with the manually
provided similarity judgements. Correspondingly, we evaluate the
performance of the probing approach by computing the Spearman
correlation between \(|| Mr_{s_i} - Mr_{s_j} ||_2\) and \(l_{i,j}\).

The number of columns of \(M\) is equal to the dimensionality of the embedding space, but we can control 
the number of rows and thus vary the dimensionality of the projection subspace. In all experiments, we use 
the sequence of the powers of two from \(2^1\) to \(2^9\), augmented with the embedding dimension of the model
(e.g., 768 for BERT base).

We  apply a similar approach for a subset of data from the \textbf{TE} task: for sentence triplets
where we have both an entailment \(e_{i}\) and a contradiction \(c_{i}\) for a given premise \(p_{i}\),
we define \(q_{i} = || Mr_{p_i} - Mr_{e_j} ||_2\) and \(r_{i} = || Mr_{p_i} - Mr_{c_j} ||_2\)
and minimise 
\begin{equation}
    \mathcal{L}_{\text{TE-triplet}} = [q_i - r_i]_{+}
\end{equation}
where \([\cdot]\) stands for \(\max(0, \cdot)\). In this manner, we encourage premises to be closer 
to their entailments than to contradictions.

By replacing Euclidean distances with cosine similarities, we can further tackle any premise--hypothesis
pair \((p_i, h_i)\) by minimising \(\mathcal{L}_{\text{TE-pair}} =\)
\begin{equation}
    \begin{cases}
        (1 - \operatorname{Cos}(Mp_i, Mh_i))^2 & \text{if } l_i = \text{entail.},\\
        (-1 - \operatorname{Cos}(Mp_i, Mh_i))^2 & \text{if } l_i = \text{contr.}
    \end{cases}
\end{equation}
where \(\operatorname{Cos}\) is cosine similarity. In this manner, we induce
 entailments to show positive cosine similarities to their premises,
and contradictions to show negative similarities to their premises.\footnote{Even though Cosine, as a symmetrical measure, is not an ideal match for asymmetrical entailment, it works well in practice \citep{reimers2019sentencebert}.}

We evaluate the TE models using accuracy. In the triplet setting, 
we count as hits all cases where \(r_i - q_i < 0\). In the sentence-pair setting, 
we follow the intuition above and consider answers to be correct if 
\(\operatorname{Cos}(Mp_i, Mh_i) > 0\) for entailments and 
\(\operatorname{Cos}(Mp_i, Mh_i) \le 0\) for contradictions.

It must be stressed that by \enquote{evaluation} we mean a proxy
measure of the informativeness of vanilla embeddings and not a
measure of how well the models can solve the original task. The labels
of the tasks themselves constitute an \enquote{abuse of notation} as
in all cases we are dealing with reformulations of the original tasks,
which in the case of NLI involve a considerable simplification. Thus,
numbers should not be compared to results on the original tasks.
Nevertheless, we believe that our proxy tasks can provide interesting
insights into the models.





\paragraph{Representation extraction}
\label{ssec:representation-extraction}

The extraction of sentence representations depends on the architecture of the model. When working 
with encoder-only MLMs, such as BERT, we follow the standard practice of averaging all token 
representations in a given layer. When working with T5-type models, which have both an encoder and a 
decoder, we hypothesise that the heavy lifting in representation learning is being done by the encoder 
and apply the same approach to it \citep[cf.\ ][]{ni-etal-2022-large}.
For causal LM models, such as GPT-2, Llama, and OPT, we extract
the representation of the last token of the input sentence.

Recall that since our goal is to probe general models, we always work
with vanilla pre-trained versions with no fine-tuning. Our structural
probing approach is, however, also applicable to fine-tuned models.

\section{Models, data, and experimental setup}
\label{sec:experiments}

\subsection{Models}\label{ssec:models}

We experiment with the following models:

\begin{itemize}
    \item \textbf{MLMs}: BERT \citep{devlin-etal-2019-bert}, RoBERTa \citep{liu2020roberta}, and
    ELECTRA \citep{clark2020electra}.
    \item \textbf{Text-to-text}: the original T5 series of models \citep{raffel2020t5} and the
    T5-efficient model series \citep{tay2022scale}.
    \item \textbf{Causal LMs}: GPT-2 \citep{radford2019language}, Llama\footnote{\url{https://huggingface.co/openlm-research}}
     and OPT \citep{zhang2022opt}.\footnote{All model checkpoints were downloaded from HuggingFace.}
\end{itemize}
We aim at providing maximum coverage by model type and size for all tasks,
but due to very high computational costs of running larger models
on large datasets (T5 XXL and causal LMs with 7b+ parameters),
even in inference mode, gaps remain.

\subsection{Datasets}\label{ssec:datasets}

For the STS task, we use the STS benchmark \citep{cer-etal-2017-semeval} distributed with
the \texttt{sentence-transformers} Python library.\footnote{\url{https://sbert.net/datasets/stsbenchmark.tsv.gz}} 

For the TE task, we use SNLI \citep{bowman-etal-2015-large}, 
MNLI \citep{williams-etal-2018-broad}, and ANLI \citep{nie-etal-2020-adversarial},
all distributed by HuggingFace. 

See the Appendix for the sizes of data splits.



\subsection{Experimental setup}\label{ssec:experiments}

All experiments are implemented using PyTorch and the \texttt{transformers} 
library \citep{wolf-etal-2020-transformers}.\footnote{Scripts for conducting 
the analyses can be found at \url{https://github.com/macleginn/semantic-subspaces-code}}
Projection matrices are implemented as single PyTorch linear layers without bias 
and are fit to data using AdamW \citep{loshchilov2019adamw} and the learning rate of \(10^{-5}\).
A~separate matrix is fitted for each combination of the model, layer, and subspace 
dimensionality. 

For the STS task, where the dataset is comparatively small, the optimisation is allowed to run for the 
maximum of 300 epochs with early stopping after 10 
epochs without improvement on the development set. For the TE task, where training
takes much longer, each optimisation is run for 5 epochs with the best checkpoint selected by 
performance on the development set. All the results are reported  for the test set. 
In the context of the STS task, we conduct 10 runs of each experiment to assess the robustness of 
results to random initialisation of projection matrices. 

\section{Results}\label{sec:results}

In this section, we first describe our presentation of the results of the experiments
(\S~\ref{ssec:presentation}) and then go over individual tasks and model types 
(\S~\ref{ssec:results-tasks}).

\subsection{Presentation of results}\label{ssec:presentation}

The result for each experiment run is a matrix with rows
corresponding to projection-subspace dimensionalities and columns corresponding to layers. When it was feasible to run the experiments several times, we obtain a matrix of averages and 
a matrix of standard deviations.
As we show below, these matrices exhibit interesting patterns of how semantic information is distributed in the models.

However, it is unwieldy to operate with a large number of such
matrices, and for summary comparison of model architectures and sizes,
we collapse them into by-layer and by-dimensionality vectors by
applying the maximum function to the columns or the rows of the
matrices respectively. In order to compare models of different sizes,
which have different numbers of layers, we further map layer numbers
to the interval [0, 1], encoding \textit{relative layer position},
such that 0 corresponds to the embedding layer and 1 to the final layer,
respectively. We visualise these results as line graphs.


\subsection{Results by task}\label{ssec:results-tasks}

\subsubsection{Semantic textual similarity}\label{sssec:results-sts}

\paragraph{Results across model architectures} 
We first assess to what extent sentence representations 
computed by encoder-only, text-to-text, and causal language models
contain subspaces where distances between representations mirror their
semantic distances according to human annotations. Here, and in the
other experiments, our results provide a lower bound on the amount of
structure and informativeness of the embeddings extracted from vanilla
pre-trained models. Assuming, however, that our probe provides a reasonable
proxy for the informativeness of the embeddings, we can also ask which
layer provides the richest embeddings and what is the minimal necessary
dimensionality of the projection subspace to achieve good results.

\begin{table}[t]
\centering
\resizebox{0.48\textwidth}{!}{
\begin{tabular}{
>{\columncolor[HTML]{FFFFFF}}r rrrrrrrrrrrr}
\multicolumn{13}{c}{\cellcolor[HTML]{FFFFFF}\textbf{BERT large cased}} \\
\multicolumn{1}{l}{\cellcolor[HTML]{FFFFFF}} &
  \cellcolor[HTML]{FFFFFF}\textbf{2} &
  \cellcolor[HTML]{FFFFFF}\textbf{4} &
  \cellcolor[HTML]{FFFFFF}\textbf{6} &
  \cellcolor[HTML]{FFFFFF}\textbf{8} &
  \cellcolor[HTML]{FFFFFF}\textbf{10} &
  \cellcolor[HTML]{FFFFFF}\textbf{12} &
  \cellcolor[HTML]{FFFFFF}\textbf{14} &
  \cellcolor[HTML]{FFFFFF}\textbf{16} &
  \cellcolor[HTML]{FFFFFF}\textbf{18} &
  \cellcolor[HTML]{FFFFFF}\textbf{20} &
  \cellcolor[HTML]{FFFFFF}\textbf{22} &
  \cellcolor[HTML]{FFFFFF}\textbf{24} \\ \cline{2-13} 
\multicolumn{1}{r|}{\cellcolor[HTML]{FFFFFF}\textbf{2}} &
  \cellcolor[HTML]{FAE1DF}0.36 &
  \cellcolor[HTML]{FCECEB}0.32 &
  \cellcolor[HTML]{FDF5F4}0.29 &
  \cellcolor[HTML]{FDF1F0}0.3 &
  \cellcolor[HTML]{FBE6E4}0.34 &
  \cellcolor[HTML]{FCEEED}0.31 &
  \cellcolor[HTML]{FEF7F6}0.28 &
  \cellcolor[HTML]{FFFDFD}0.26 &
  \cellcolor[HTML]{FEF8F8}0.28 &
  \cellcolor[HTML]{FAE1DE}0.36 &
  \cellcolor[HTML]{FEF8F8}0.28 &
  \cellcolor[HTML]{FFFFFF}0.25 \\
\multicolumn{1}{r|}{\cellcolor[HTML]{FFFFFF}\textbf{4}} &
  \cellcolor[HTML]{F3BCB7}0.49 &
  \cellcolor[HTML]{F7D4D1}0.4 &
  \cellcolor[HTML]{F8D8D5}0.39 &
  \cellcolor[HTML]{F7D4D1}0.41 &
  \cellcolor[HTML]{F3C0BB}0.48 &
  \cellcolor[HTML]{F4C3BE}0.47 &
  \cellcolor[HTML]{F8D9D6}0.39 &
  \cellcolor[HTML]{F6CFCB}0.42 &
  \cellcolor[HTML]{F1B2AD}0.52 &
  \cellcolor[HTML]{F2B7B2}0.51 &
  \cellcolor[HTML]{F6D0CC}0.42 &
  \cellcolor[HTML]{FAE5E3}0.35 \\
\multicolumn{1}{r|}{\cellcolor[HTML]{FFFFFF}\textbf{8}} &
  \cellcolor[HTML]{EEA49E}0.57 &
  \cellcolor[HTML]{EEA49E}0.57 &
  \cellcolor[HTML]{EEA49E}0.57 &
  \cellcolor[HTML]{F0AFA9}0.54 &
  \cellcolor[HTML]{EDA19A}0.59 &
  \cellcolor[HTML]{EC9B94}0.61 &
  \cellcolor[HTML]{EC9C95}0.6 &
  \cellcolor[HTML]{ED9E97}0.6 &
  \cellcolor[HTML]{EB958E}0.63 &
  \cellcolor[HTML]{EC9992}0.61 &
  \cellcolor[HTML]{EC9790}0.62 &
  \cellcolor[HTML]{EB958E}0.63 \\
\multicolumn{1}{r|}{\cellcolor[HTML]{FFFFFF}\textbf{16}} &
  \cellcolor[HTML]{EB948C}0.63 &
  \cellcolor[HTML]{EB958E}0.63 &
  \cellcolor[HTML]{EB948D}0.63 &
  \cellcolor[HTML]{EA9189}0.64 &
  \cellcolor[HTML]{E98C84}0.66 &
  \cellcolor[HTML]{EA8D85}0.66 &
  \cellcolor[HTML]{EA8F87}0.65 &
  \cellcolor[HTML]{EA8E86}0.65 &
  \cellcolor[HTML]{E98981}0.67 &
  \cellcolor[HTML]{E98981}0.67 &
  \cellcolor[HTML]{E98A81}0.67 &
  \cellcolor[HTML]{E8867E}0.68 \\
\multicolumn{1}{r|}{\cellcolor[HTML]{FFFFFF}\textbf{32}} &
  \cellcolor[HTML]{E98C84}0.66 &
  \cellcolor[HTML]{EA8F87}0.65 &
  \cellcolor[HTML]{EA8E86}0.65 &
  \cellcolor[HTML]{E98A82}0.67 &
  \cellcolor[HTML]{E8857D}0.68 &
  \cellcolor[HTML]{E8847B}0.69 &
  \cellcolor[HTML]{E9877F}0.68 &
  \cellcolor[HTML]{E8867E}0.68 &
  \cellcolor[HTML]{E8827A}0.69 &
  \cellcolor[HTML]{E8837B}\textbf{0.69} &
  \cellcolor[HTML]{E78078}\textbf{0.7} &
  \cellcolor[HTML]{E78078}\textbf{0.7} \\
\multicolumn{1}{r|}{\cellcolor[HTML]{FFFFFF}\textbf{64}} &
  \cellcolor[HTML]{E98981}0.67 &
  \cellcolor[HTML]{E98A82}0.67 &
  \cellcolor[HTML]{E98B83}0.66 &
  \cellcolor[HTML]{E8877E}0.68 &
  \cellcolor[HTML]{E78279}0.7 &
  \cellcolor[HTML]{E78178}0.7 &
  \cellcolor[HTML]{E8827A}0.69 &
  \cellcolor[HTML]{E8837A}0.69 &
  \cellcolor[HTML]{E8847B}0.69 &
  \cellcolor[HTML]{E8837B}\textbf{0.69} &
  \cellcolor[HTML]{E77F76}\textbf{0.71} &
  \cellcolor[HTML]{E77D74}\textbf{0.71} \\
\multicolumn{1}{r|}{\cellcolor[HTML]{FFFFFF}\textbf{128}} &
  \cellcolor[HTML]{EEA59F}0.57 &
  \cellcolor[HTML]{E98A81}0.67 &
  \cellcolor[HTML]{E98981}0.67 &
  \cellcolor[HTML]{E8867E}0.68 &
  \cellcolor[HTML]{E8857D}0.68 &
  \cellcolor[HTML]{E77F77}0.7 &
  \cellcolor[HTML]{E88279}0.7 &
  \cellcolor[HTML]{EFAAA4}0.55 &
  \cellcolor[HTML]{F2B7B2}0.51 &
  \cellcolor[HTML]{F3BDB8}0.49 &
  \cellcolor[HTML]{F0AEA9}0.54 &
  \cellcolor[HTML]{EEA49E}0.57 \\
\multicolumn{1}{r|}{\cellcolor[HTML]{FFFFFF}\textbf{256}} &
  \cellcolor[HTML]{EEA69F}0.57 &
  \cellcolor[HTML]{EFA9A3}0.56 &
  \cellcolor[HTML]{EFAAA4}0.55 &
  \cellcolor[HTML]{EFABA6}0.55 &
  \cellcolor[HTML]{EFA7A1}0.56 &
  \cellcolor[HTML]{EFA8A2}0.56 &
  \cellcolor[HTML]{F1B1AC}0.53 &
  \cellcolor[HTML]{F2B9B4}0.5 &
  \cellcolor[HTML]{F2BAB5}0.5 &
  \cellcolor[HTML]{F3BFBB}0.48 &
  \cellcolor[HTML]{F2B9B4}0.5 &
  \cellcolor[HTML]{EFA7A1}0.56 \\
\multicolumn{1}{r|}{\cellcolor[HTML]{FFFFFF}\textbf{512}} &
  \cellcolor[HTML]{EEA6A0}0.57 &
  \cellcolor[HTML]{EFAAA4}0.55 &
  \cellcolor[HTML]{F0ADA7}0.54 &
  \cellcolor[HTML]{F0AEA8}0.54 &
  \cellcolor[HTML]{EFA8A2}0.56 &
  \cellcolor[HTML]{EFAAA4}0.55 &
  \cellcolor[HTML]{F1B2AD}0.53 &
  \cellcolor[HTML]{F2B9B4}0.5 &
  \cellcolor[HTML]{F2BBB6}0.49 &
  \cellcolor[HTML]{F3C0BB}0.48 &
  \cellcolor[HTML]{F2BAB5}0.5 &
  \cellcolor[HTML]{EFA7A1}0.56 \\
\multicolumn{1}{r|}{\cellcolor[HTML]{FFFFFF}\textbf{1024}} &
  \cellcolor[HTML]{EEA6A0}0.57 &
  \cellcolor[HTML]{EFAAA4}0.55 &
  \cellcolor[HTML]{F0ADA8}0.54 &
  \cellcolor[HTML]{F0AEA9}0.54 &
  \cellcolor[HTML]{EFA9A3}0.56 &
  \cellcolor[HTML]{EFAAA4}0.55 &
  \cellcolor[HTML]{F1B3AD}0.52 &
  \cellcolor[HTML]{F2BAB5}0.5 &
  \cellcolor[HTML]{F3BCB7}0.49 &
  \cellcolor[HTML]{F4C2BD}0.47 &
  \cellcolor[HTML]{F2BBB6}0.49 &
  \cellcolor[HTML]{EFA8A2}0.56 \\
\multicolumn{1}{r|}{\cellcolor[HTML]{FFFFFF}\textbf{1024$'$}} &
  \cellcolor[HTML]{EEA6A0}0.57 &
  \cellcolor[HTML]{EFACA6}0.55 &
  \cellcolor[HTML]{F0AFAA}0.54 &
  \cellcolor[HTML]{F0B0AA}0.53 &
  \cellcolor[HTML]{EEA6A0}0.57 &
  \cellcolor[HTML]{EFAAA5}0.55 &
  \cellcolor[HTML]{F1B2AD}0.52 &
  \cellcolor[HTML]{F2B8B3}0.5 &
  \cellcolor[HTML]{F2B6B1}0.51 &
  \cellcolor[HTML]{F3BDB8}0.49 &
  \cellcolor[HTML]{F2B9B4}0.5 &
  \cellcolor[HTML]{EEA49D}0.58 \\
\multicolumn{13}{c}{\cellcolor[HTML]{FFFFFF}\textbf{RoBERTa   large}} \\
\multicolumn{1}{r|}{\cellcolor[HTML]{FFFFFF}\textbf{2}} &
  \cellcolor[HTML]{FAE5E3}0.31 &
  \cellcolor[HTML]{FCEFEE}0.27 &
  \cellcolor[HTML]{F9DBD9}0.35 &
  \cellcolor[HTML]{F9DCD9}0.35 &
  \cellcolor[HTML]{FAE5E3}0.31 &
  \cellcolor[HTML]{FCECEB}0.28 &
  \cellcolor[HTML]{F8D7D4}0.37 &
  \cellcolor[HTML]{FAE4E2}0.31 &
  \cellcolor[HTML]{FBEAE8}0.29 &
  \cellcolor[HTML]{FDF0EF}0.26 &
  \cellcolor[HTML]{FCEEED}0.27 &
  \cellcolor[HTML]{FFFFFF}0.19 \\
\multicolumn{1}{r|}{\cellcolor[HTML]{FFFFFF}\textbf{4}} &
  \cellcolor[HTML]{F6CFCC}0.41 &
  \cellcolor[HTML]{F6CFCC}0.41 &
  \cellcolor[HTML]{F1B6B1}0.52 &
  \cellcolor[HTML]{F5C7C4}0.44 &
  \cellcolor[HTML]{F8D7D4}0.37 &
  \cellcolor[HTML]{F8D7D5}0.37 &
  \cellcolor[HTML]{F6CDCA}0.41 &
  \cellcolor[HTML]{F6D0CD}0.4 &
  \cellcolor[HTML]{F8D6D3}0.38 &
  \cellcolor[HTML]{F7D5D2}0.38 &
  \cellcolor[HTML]{F8D7D4}0.37 &
  \cellcolor[HTML]{FDF3F2}0.25 \\
\multicolumn{1}{r|}{\cellcolor[HTML]{FFFFFF}\textbf{8}} &
  \cellcolor[HTML]{F0ADA7}0.56 &
  \cellcolor[HTML]{EFA8A2}0.58 &
  \cellcolor[HTML]{EEA19B}0.61 &
  \cellcolor[HTML]{ED9D97}0.63 &
  \cellcolor[HTML]{ED9F98}0.62 &
  \cellcolor[HTML]{ED9D97}0.63 &
  \cellcolor[HTML]{EC9B94}0.64 &
  \cellcolor[HTML]{EEA29C}0.61 &
  \cellcolor[HTML]{F3BFBA}0.48 &
  \cellcolor[HTML]{F1B2AC}0.54 &
  \cellcolor[HTML]{F3BEBA}0.48 &
  \cellcolor[HTML]{FBE6E4}0.3 \\
\multicolumn{1}{r|}{\cellcolor[HTML]{FFFFFF}\textbf{16}} &
  \cellcolor[HTML]{EEA29C}0.61 &
  \cellcolor[HTML]{EC9B94}0.64 &
  \cellcolor[HTML]{EB958E}0.66 &
  \cellcolor[HTML]{EB928A}0.68 &
  \cellcolor[HTML]{EA8E86}0.7 &
  \cellcolor[HTML]{E98A82}0.72 &
  \cellcolor[HTML]{E98981}0.72 &
  \cellcolor[HTML]{E98980}0.72 &
  \cellcolor[HTML]{E8867D}0.73 &
  \cellcolor[HTML]{E8877E}0.73 &
  \cellcolor[HTML]{E98880}0.72 &
  \cellcolor[HTML]{F8D9D6}0.36 \\
\multicolumn{1}{r|}{\cellcolor[HTML]{FFFFFF}\textbf{32}} &
  \cellcolor[HTML]{EC9A93}0.64 &
  \cellcolor[HTML]{EB938B}0.68 &
  \cellcolor[HTML]{EA8E86}0.7 &
  \cellcolor[HTML]{E98B83}0.71 &
  \cellcolor[HTML]{E8867E}0.73 &
  \cellcolor[HTML]{E8837B}0.74 &
  \cellcolor[HTML]{E8847B}0.74 &
  \cellcolor[HTML]{E78178}0.75 &
  \cellcolor[HTML]{E78077}0.76 &
  \cellcolor[HTML]{E78178}0.76 &
  \cellcolor[HTML]{E88279}0.75 &
  \cellcolor[HTML]{F6CBC8}0.42 \\
\multicolumn{1}{r|}{\cellcolor[HTML]{FFFFFF}\textbf{64}} &
  \cellcolor[HTML]{EB978F}0.66 &
  \cellcolor[HTML]{EA9089}0.69 &
  \cellcolor[HTML]{E98C84}0.71 &
  \cellcolor[HTML]{E98981}0.72 &
  \cellcolor[HTML]{E8837A}0.75 &
  \cellcolor[HTML]{E78178}0.76 &
  \cellcolor[HTML]{E78077}0.76 &
  \cellcolor[HTML]{E78077}0.76 &
  \cellcolor[HTML]{E77E75}\textbf{0.77} &
  \cellcolor[HTML]{E77E75}\textbf{0.77} &
  \cellcolor[HTML]{E77F76}\textbf{0.77} &
  \cellcolor[HTML]{E98981}0.72 \\
\multicolumn{1}{r|}{\cellcolor[HTML]{FFFFFF}\textbf{128}} &
  \cellcolor[HTML]{EFA8A2}0.58 &
  \cellcolor[HTML]{F0AEA9}0.55 &
  \cellcolor[HTML]{EFABA5}0.57 &
  \cellcolor[HTML]{F0ACA6}0.56 &
  \cellcolor[HTML]{E8837A}0.75 &
  \cellcolor[HTML]{E78279}0.75 &
  \cellcolor[HTML]{EFA9A3}0.58 &
  \cellcolor[HTML]{E8837B}0.74 &
  \cellcolor[HTML]{E77E75}\textbf{0.77} &
  \cellcolor[HTML]{E77E75}\textbf{0.77} &
  \cellcolor[HTML]{E77E75}\textbf{0.77} &
  \cellcolor[HTML]{E8867E}0.73 \\
\multicolumn{1}{r|}{\cellcolor[HTML]{FFFFFF}\textbf{256}} &
  \cellcolor[HTML]{F1B2AD}0.54 &
  \cellcolor[HTML]{F0B1AB}0.54 &
  \cellcolor[HTML]{F0ADA7}0.56 &
  \cellcolor[HTML]{F0AEA8}0.55 &
  \cellcolor[HTML]{EFABA5}0.57 &
  \cellcolor[HTML]{F0ACA6}0.56 &
  \cellcolor[HTML]{F0AEA9}0.55 &
  \cellcolor[HTML]{F0AEA9}0.55 &
  \cellcolor[HTML]{F0ACA6}0.56 &
  \cellcolor[HTML]{EFAAA4}0.57 &
  \cellcolor[HTML]{EFA8A2}0.58 &
  \cellcolor[HTML]{E8847B}0.74 \\
\multicolumn{1}{r|}{\cellcolor[HTML]{FFFFFF}\textbf{512}} &
  \cellcolor[HTML]{F1B2AD}0.54 &
  \cellcolor[HTML]{F1B1AC}0.54 &
  \cellcolor[HTML]{F0ADA7}0.56 &
  \cellcolor[HTML]{F0B0AA}0.55 &
  \cellcolor[HTML]{F0AEA8}0.55 &
  \cellcolor[HTML]{F0AEA8}0.55 &
  \cellcolor[HTML]{F0B1AB}0.54 &
  \cellcolor[HTML]{F1B1AC}0.54 &
  \cellcolor[HTML]{F0AFA9}0.55 &
  \cellcolor[HTML]{F0ADA7}0.56 &
  \cellcolor[HTML]{EFABA5}0.57 &
  \cellcolor[HTML]{E8837B}0.74 \\
\multicolumn{1}{r|}{\cellcolor[HTML]{FFFFFF}\textbf{1024}} &
  \cellcolor[HTML]{F1B2AD}0.54 &
  \cellcolor[HTML]{F1B2AC}0.54 &
  \cellcolor[HTML]{F0ADA8}0.56 &
  \cellcolor[HTML]{F0B0AB}0.54 &
  \cellcolor[HTML]{F0AFA9}0.55 &
  \cellcolor[HTML]{F0B0AA}0.55 &
  \cellcolor[HTML]{F1B2AC}0.54 &
  \cellcolor[HTML]{F1B2AD}0.54 &
  \cellcolor[HTML]{F0B0AB}0.54 &
  \cellcolor[HTML]{F0AEA9}0.55 &
  \cellcolor[HTML]{F0ADA7}0.56 &
  \cellcolor[HTML]{F1B6B1}0.52 \\
\multicolumn{1}{r|}{\cellcolor[HTML]{FFFFFF}\textbf{1024$'$}} &
  \cellcolor[HTML]{F1B4AE}0.53 &
  \cellcolor[HTML]{F1B4AF}0.53 &
  \cellcolor[HTML]{EFABA5}0.57 &
  \cellcolor[HTML]{EFAAA4}0.57 &
  \cellcolor[HTML]{EFAAA4}0.57 &
  \cellcolor[HTML]{EFAAA4}0.57 &
  \cellcolor[HTML]{EFA9A3}0.58 &
  \cellcolor[HTML]{EFA9A3}0.58 &
  \cellcolor[HTML]{EFA9A3}0.57 &
  \cellcolor[HTML]{EFA8A2}0.58 &
  \cellcolor[HTML]{EFA9A3}0.58 &
  \cellcolor[HTML]{F3C0BC}0.47 \\
\multicolumn{13}{c}{\cellcolor[HTML]{FFFFFF}\textbf{ELECTRA large}} \\
\multicolumn{1}{r|}{\cellcolor[HTML]{FFFFFF}\textbf{2}} &
  \cellcolor[HTML]{FBE9E7}0.29 &
  \cellcolor[HTML]{FAE2E0}0.31 &
  \cellcolor[HTML]{F9DEDC}0.33 &
  \cellcolor[HTML]{F9DCD9}0.34 &
  \cellcolor[HTML]{FAE4E2}0.3 &
  \cellcolor[HTML]{FBEAE9}0.28 &
  \cellcolor[HTML]{F8D8D6}0.35 &
  \cellcolor[HTML]{F8D6D4}0.36 &
  \cellcolor[HTML]{F9DFDC}0.33 &
  \cellcolor[HTML]{FDF2F1}0.24 &
  \cellcolor[HTML]{FEF6F6}0.23 &
  \cellcolor[HTML]{FFFFFF}0.19 \\
\multicolumn{1}{r|}{\cellcolor[HTML]{FFFFFF}\textbf{4}} &
  \cellcolor[HTML]{F7D5D2}0.37 &
  \cellcolor[HTML]{F6CBC8}0.41 &
  \cellcolor[HTML]{F3BEB9}0.46 &
  \cellcolor[HTML]{F3BCB7}0.47 &
  \cellcolor[HTML]{F6CECB}0.4 &
  \cellcolor[HTML]{F7D1CE}0.38 &
  \cellcolor[HTML]{F6CDC9}0.4 &
  \cellcolor[HTML]{F4C3BF}0.44 &
  \cellcolor[HTML]{F4C2BE}0.45 &
  \cellcolor[HTML]{FBE8E6}0.29 &
  \cellcolor[HTML]{FEF6F5}0.23 &
  \cellcolor[HTML]{FFFEFE}0.2 \\
\multicolumn{1}{r|}{\cellcolor[HTML]{FFFFFF}\textbf{8}} &
  \cellcolor[HTML]{F5C9C6}0.42 &
  \cellcolor[HTML]{F0ADA7}0.54 &
  \cellcolor[HTML]{EC9A93}0.62 &
  \cellcolor[HTML]{EFABA5}0.55 &
  \cellcolor[HTML]{EB968F}0.63 &
  \cellcolor[HTML]{F3BCB7}0.47 &
  \cellcolor[HTML]{EC9891}0.62 &
  \cellcolor[HTML]{EB958E}0.63 &
  \cellcolor[HTML]{EB948D}0.64 &
  \cellcolor[HTML]{F3BEBA}0.46 &
  \cellcolor[HTML]{FAE1DF}0.32 &
  \cellcolor[HTML]{FEF7F6}0.23 \\
\multicolumn{1}{r|}{\cellcolor[HTML]{FFFFFF}\textbf{16}} &
  \cellcolor[HTML]{ED9F99}0.59 &
  \cellcolor[HTML]{E98A82}0.68 &
  \cellcolor[HTML]{E98880}0.69 &
  \cellcolor[HTML]{E8877E}0.7 &
  \cellcolor[HTML]{E98880}0.69 &
  \cellcolor[HTML]{E9887F}0.69 &
  \cellcolor[HTML]{E9877F}0.69 &
  \cellcolor[HTML]{E98981}0.69 &
  \cellcolor[HTML]{E98A82}0.68 &
  \cellcolor[HTML]{EA8D85}0.67 &
  \cellcolor[HTML]{ED9D96}0.6 &
  \cellcolor[HTML]{F8D7D4}0.36 \\
\multicolumn{1}{r|}{\cellcolor[HTML]{FFFFFF}\textbf{32}} &
  \cellcolor[HTML]{E98A82}0.68 &
  \cellcolor[HTML]{E8837B}0.71 &
  \cellcolor[HTML]{E8827A}0.71 &
  \cellcolor[HTML]{E78179}0.72 &
  \cellcolor[HTML]{E88279}0.72 &
  \cellcolor[HTML]{E8827A}0.71 &
  \cellcolor[HTML]{E88279}0.72 &
  \cellcolor[HTML]{E8847C}0.71 &
  \cellcolor[HTML]{E8857C}0.7 &
  \cellcolor[HTML]{E98981}0.69 &
  \cellcolor[HTML]{EB958E}0.63 &
  \cellcolor[HTML]{F2BBB6}0.48 \\
\multicolumn{1}{r|}{\cellcolor[HTML]{FFFFFF}\textbf{64}} &
  \cellcolor[HTML]{E8877F}0.69 &
  \cellcolor[HTML]{E78179}0.72 &
  \cellcolor[HTML]{E78077}0.72 &
  \cellcolor[HTML]{E77E75}\textbf{0.73} &
  \cellcolor[HTML]{E77E76}\textbf{0.73} &
  \cellcolor[HTML]{E78178}0.72 &
  \cellcolor[HTML]{E88279}0.72 &
  \cellcolor[HTML]{E8837A}0.71 &
  \cellcolor[HTML]{E8857C}0.7 &
  \cellcolor[HTML]{E9877F}0.69 &
  \cellcolor[HTML]{EB938C}0.64 &
  \cellcolor[HTML]{F2B8B3}0.49 \\
\multicolumn{1}{r|}{\cellcolor[HTML]{FFFFFF}\textbf{128}} &
  \cellcolor[HTML]{E8857D}0.7 &
  \cellcolor[HTML]{E77F76}\textbf{0.73} &
  \cellcolor[HTML]{E77E75}\textbf{0.73} &
  \cellcolor[HTML]{E77D74}\textbf{0.74} &
  \cellcolor[HTML]{E77E75}\textbf{0.73} &
  \cellcolor[HTML]{E78178}0.72 &
  \cellcolor[HTML]{E8837A}0.71 &
  \cellcolor[HTML]{E9877F}0.69 &
  \cellcolor[HTML]{E8867D}0.7 &
  \cellcolor[HTML]{E98981}0.69 &
  \cellcolor[HTML]{EB948C}0.64 &
  \cellcolor[HTML]{F2B7B2}0.49 \\
\multicolumn{1}{r|}{\cellcolor[HTML]{FFFFFF}\textbf{256}} &
  \cellcolor[HTML]{E8847B}0.71 &
  \cellcolor[HTML]{EB968F}0.63 &
  \cellcolor[HTML]{EC9A93}0.62 &
  \cellcolor[HTML]{E77E75}\textbf{0.73} &
  \cellcolor[HTML]{EEA29C}0.58 &
  \cellcolor[HTML]{F0AEA8}0.53 &
  \cellcolor[HTML]{F0B0AA}0.52 &
  \cellcolor[HTML]{F0B0AA}0.52 &
  \cellcolor[HTML]{F1B5B0}0.5 &
  \cellcolor[HTML]{F3C0BC}0.46 &
  \cellcolor[HTML]{EB958E}0.63 &
  \cellcolor[HTML]{F1B6B1}0.5 \\
\multicolumn{1}{r|}{\cellcolor[HTML]{FFFFFF}\textbf{512}} &
  \cellcolor[HTML]{ED9F98}0.59 &
  \cellcolor[HTML]{EC9C95}0.61 &
  \cellcolor[HTML]{ED9C95}0.61 &
  \cellcolor[HTML]{EDA19A}0.59 &
  \cellcolor[HTML]{EFA9A3}0.55 &
  \cellcolor[HTML]{F0B0AB}0.52 &
  \cellcolor[HTML]{F1B2AC}0.51 &
  \cellcolor[HTML]{F1B2AD}0.51 &
  \cellcolor[HTML]{F2B8B3}0.49 &
  \cellcolor[HTML]{F4C2BE}0.45 &
  \cellcolor[HTML]{F7D1CE}0.38 &
  \cellcolor[HTML]{F2B7B2}0.49 \\
\multicolumn{1}{r|}{\cellcolor[HTML]{FFFFFF}\textbf{1024}} &
  \cellcolor[HTML]{EDA19A}0.59 &
  \cellcolor[HTML]{ED9C96}0.6 &
  \cellcolor[HTML]{ED9D96}0.6 &
  \cellcolor[HTML]{EEA29C}0.58 &
  \cellcolor[HTML]{EFAAA5}0.55 &
  \cellcolor[HTML]{F1B1AC}0.52 &
  \cellcolor[HTML]{F1B4AE}0.51 &
  \cellcolor[HTML]{F1B3AE}0.51 &
  \cellcolor[HTML]{F2B9B4}0.48 &
  \cellcolor[HTML]{F4C4C0}0.44 &
  \cellcolor[HTML]{F7D3D0}0.38 &
  \cellcolor[HTML]{FAE3E1}0.31 \\
\multicolumn{1}{r|}{\cellcolor[HTML]{FFFFFF}\textbf{1024$'$}} &
  \cellcolor[HTML]{EEA59E}0.57 &
  \cellcolor[HTML]{ED9F98}0.59 &
  \cellcolor[HTML]{ED9E97}0.6 &
  \cellcolor[HTML]{EEA49E}0.57 &
  \cellcolor[HTML]{EFACA6}0.54 &
  \cellcolor[HTML]{F0AFA9}0.53 &
  \cellcolor[HTML]{F0ADA7}0.54 &
  \cellcolor[HTML]{F0ACA6}0.54 &
  \cellcolor[HTML]{F1B3AE}0.51 &
  \cellcolor[HTML]{F4C5C1}0.44 &
  \cellcolor[HTML]{F8DAD8}0.35 &
  \cellcolor[HTML]{FEF6F5}0.23 \\
\multicolumn{13}{c}{\cellcolor[HTML]{FFFFFF}\textbf{T5 large}} \\
\multicolumn{1}{r|}{\cellcolor[HTML]{FFFFFF}\textbf{2}} &
  \cellcolor[HTML]{F9E0DE}0.27 &
  \cellcolor[HTML]{F9DDDB}0.28 &
  \cellcolor[HTML]{F9DFDC}0.28 &
  \cellcolor[HTML]{FAE4E2}0.25 &
  \cellcolor[HTML]{FAE4E2}0.25 &
  \cellcolor[HTML]{FAE4E2}0.25 &
  \cellcolor[HTML]{FBE7E5}0.24 &
  \cellcolor[HTML]{FBE8E7}0.23 &
  \cellcolor[HTML]{FCEEED}0.21 &
  \cellcolor[HTML]{FCEEED}0.21 &
  \cellcolor[HTML]{FEF7F7}0.16 &
  \cellcolor[HTML]{FCEFEE}0.2 \\
\multicolumn{1}{r|}{\cellcolor[HTML]{FFFFFF}\textbf{4}} &
  \cellcolor[HTML]{F6CBC8}0.37 &
  \cellcolor[HTML]{F6CDC9}0.36 &
  \cellcolor[HTML]{F6CDC9}0.36 &
  \cellcolor[HTML]{F7D1CE}0.34 &
  \cellcolor[HTML]{F6CECB}0.36 &
  \cellcolor[HTML]{F7D3D0}0.33 &
  \cellcolor[HTML]{F7D3D0}0.33 &
  \cellcolor[HTML]{F9DBD9}0.29 &
  \cellcolor[HTML]{F9DFDD}0.28 &
  \cellcolor[HTML]{FAE1DF}0.27 &
  \cellcolor[HTML]{FBEAE9}0.22 &
  \cellcolor[HTML]{FAE5E3}0.25 \\
\multicolumn{1}{r|}{\cellcolor[HTML]{FFFFFF}\textbf{8}} &
  \cellcolor[HTML]{F2BAB6}0.45 &
  \cellcolor[HTML]{F2BAB5}0.45 &
  \cellcolor[HTML]{F3BEBA}0.43 &
  \cellcolor[HTML]{F3C0BB}0.42 &
  \cellcolor[HTML]{F3C0BB}0.42 &
  \cellcolor[HTML]{F4C2BE}0.41 &
  \cellcolor[HTML]{F4C2BE}0.41 &
  \cellcolor[HTML]{F6CDCA}0.36 &
  \cellcolor[HTML]{F7D5D2}0.32 &
  \cellcolor[HTML]{F8D6D3}0.32 &
  \cellcolor[HTML]{F9DBD8}0.3 &
  \cellcolor[HTML]{F8D7D5}0.31 \\
\multicolumn{1}{r|}{\cellcolor[HTML]{FFFFFF}\textbf{16}} &
  \cellcolor[HTML]{F0B0AB}0.49 &
  \cellcolor[HTML]{F0AFAA}0.5 &
  \cellcolor[HTML]{F0B0AA}0.5 &
  \cellcolor[HTML]{F1B3AE}0.48 &
  \cellcolor[HTML]{F1B4AF}0.48 &
  \cellcolor[HTML]{F2B7B2}0.46 &
  \cellcolor[HTML]{F2BAB5}0.45 &
  \cellcolor[HTML]{F4C5C1}0.4 &
  \cellcolor[HTML]{F6CDCA}0.36 &
  \cellcolor[HTML]{F6CECB}0.36 &
  \cellcolor[HTML]{F7D4D1}0.33 &
  \cellcolor[HTML]{F6D0CC}0.35 \\
\multicolumn{1}{r|}{\cellcolor[HTML]{FFFFFF}\textbf{32}} &
  \cellcolor[HTML]{EFABA5}0.52 &
  \cellcolor[HTML]{EFA8A2}0.53 &
  \cellcolor[HTML]{EFA9A3}0.53 &
  \cellcolor[HTML]{EFABA5}0.52 &
  \cellcolor[HTML]{F0ACA7}0.51 &
  \cellcolor[HTML]{F0B0AB}0.49 &
  \cellcolor[HTML]{F1B1AC}0.49 &
  \cellcolor[HTML]{F3C0BB}0.42 &
  \cellcolor[HTML]{F5C8C4}0.39 &
  \cellcolor[HTML]{F5CBC7}0.37 &
  \cellcolor[HTML]{F6D0CD}0.35 &
  \cellcolor[HTML]{F5C6C2}0.39 \\
\multicolumn{1}{r|}{\cellcolor[HTML]{FFFFFF}\textbf{64}} &
  \cellcolor[HTML]{EEA6A0}0.54 &
  \cellcolor[HTML]{EEA29C}0.56 &
  \cellcolor[HTML]{EEA39D}0.56 &
  \cellcolor[HTML]{EEA6A0}0.54 &
  \cellcolor[HTML]{EFA8A2}0.53 &
  \cellcolor[HTML]{EFAAA4}0.53 &
  \cellcolor[HTML]{F0ACA7}0.51 &
  \cellcolor[HTML]{F2BBB7}0.44 &
  \cellcolor[HTML]{F4C5C1}0.4 &
  \cellcolor[HTML]{F5C7C3}0.39 &
  \cellcolor[HTML]{F6CCC9}0.36 &
  \cellcolor[HTML]{F2BAB5}0.45 \\
\multicolumn{1}{r|}{\cellcolor[HTML]{FFFFFF}\textbf{128}} &
  \cellcolor[HTML]{EEA49D}0.55 &
  \cellcolor[HTML]{EDA19A}0.57 &
  \cellcolor[HTML]{EDA19A}0.57 &
  \cellcolor[HTML]{EEA39D}0.56 &
  \cellcolor[HTML]{EEA59F}0.55 &
  \cellcolor[HTML]{EFA8A2}0.53 &
  \cellcolor[HTML]{EFA9A3}0.53 &
  \cellcolor[HTML]{F2BAB5}0.45 &
  \cellcolor[HTML]{F4C4C0}0.4 &
  \cellcolor[HTML]{F4C5C1}0.4 &
  \cellcolor[HTML]{F6CCC9}0.36 &
  \cellcolor[HTML]{E77D74}\textbf{0.74} \\
\multicolumn{1}{r|}{\cellcolor[HTML]{FFFFFF}\textbf{256}} &
  \cellcolor[HTML]{EEA29C}0.56 &
  \cellcolor[HTML]{ED9F99}0.57 &
  \cellcolor[HTML]{ED9F99}0.57 &
  \cellcolor[HTML]{EEA29B}0.56 &
  \cellcolor[HTML]{EEA39D}0.56 &
  \cellcolor[HTML]{EEA6A0}0.54 &
  \cellcolor[HTML]{EFA8A1}0.54 &
  \cellcolor[HTML]{F2BAB5}0.45 &
  \cellcolor[HTML]{F4C2BE}0.41 &
  \cellcolor[HTML]{F4C5C1}0.4 &
  \cellcolor[HTML]{F6CCC9}0.37 &
  \cellcolor[HTML]{E67C73}\textbf{0.74} \\
\multicolumn{1}{r|}{\cellcolor[HTML]{FFFFFF}\textbf{512}} &
  \cellcolor[HTML]{EEA29B}0.56 &
  \cellcolor[HTML]{ED9F98}0.58 &
  \cellcolor[HTML]{ED9F98}0.58 &
  \cellcolor[HTML]{EEA19B}0.57 &
  \cellcolor[HTML]{EEA39C}0.56 &
  \cellcolor[HTML]{EEA6A0}0.54 &
  \cellcolor[HTML]{EFA7A1}0.54 &
  \cellcolor[HTML]{F2B9B4}0.46 &
  \cellcolor[HTML]{F4C2BE}0.41 &
  \cellcolor[HTML]{F4C5C1}0.4 &
  \cellcolor[HTML]{F5CBC7}0.37 &
  \cellcolor[HTML]{E77D74}0.73 \\
\multicolumn{1}{r|}{\cellcolor[HTML]{FFFFFF}\textbf{1024}} &
  \cellcolor[HTML]{EEA19B}0.56 &
  \cellcolor[HTML]{ED9F98}0.58 &
  \cellcolor[HTML]{ED9F98}0.58 &
  \cellcolor[HTML]{EDA19A}0.57 &
  \cellcolor[HTML]{EEA29C}0.56 &
  \cellcolor[HTML]{EEA59F}0.55 &
  \cellcolor[HTML]{EEA6A0}0.54 &
  \cellcolor[HTML]{F2B8B3}0.46 &
  \cellcolor[HTML]{F4C2BE}0.41 &
  \cellcolor[HTML]{F4C4C0}0.4 &
  \cellcolor[HTML]{F5CBC7}0.37 &
  \cellcolor[HTML]{E77F76}0.72 \\
\multicolumn{1}{r|}{\cellcolor[HTML]{FFFFFF}\textbf{1024$'$}} &
  \cellcolor[HTML]{EFA7A1}0.54 &
  \cellcolor[HTML]{EFA7A0}0.54 &
  \cellcolor[HTML]{F0ADA7}0.51 &
  \cellcolor[HTML]{F2BBB6}0.44 &
  \cellcolor[HTML]{F4C3BF}0.41 &
  \cellcolor[HTML]{F5C9C6}0.38 &
  \cellcolor[HTML]{F8D9D6}0.31 &
  \cellcolor[HTML]{FCEBE9}0.22 &
  \cellcolor[HTML]{FDF0EF}0.2 &
  \cellcolor[HTML]{FFFBFB}0.14 &
  \cellcolor[HTML]{FFFDFD}0.14 &
  \cellcolor[HTML]{F1B2AD}0.49 \\ 
\end{tabular}
}
\caption{Spearman correlations of sentence-similarity scores derived via projection from 
averaged-token representations by model, layer (columns), and subspace dimensionality (rows)
with the STS benchmark scores. 1024$'$ stands for using vanilla 
representations without projection. The results are averaged over ten runs.}
\label{tab:sts-bert-roberta}
\end{table}

Table~\ref{tab:sts-bert-roberta} shows the full results for 
MLMs. (For space considerations, odd-numbered layers were omitted: 
they continue the same pattern.)
It can be seen that the task can be solved rather
well using only projected vanilla embeddings and that, while RoBERTa
shows better performance than BERT (\textit{r}~= 0.77 vs.\ 0.71), best
results are achieved using the same setup: extracting representations
from the layers close to the last one and projecting them to 64 or 128
dimensions. ELECTRA, whose performance is in between the classic MLMs
(\textit{r}~= 0.74) can also be made to perform well by using
128-dimensional subspace, but its best performance is achieved much
earlier in the model, on layers 8--10, and then slowly degrades.  This
demonstrates that the specialisation of higher levels on semantic
features, characteristic of BERT-like models \citep{li-etal-2021-bert}, 
does not apply straightforwardly to ELECTRA,
which raises the question of what kind of hierarchy of linguistic
features ELECTRA encodes. In line with the results by
\citet{chi-etal-2020-finding}, the use of very high-dimensional subspaces,
with or without projection, leads to bad performance.

The encoder from T5 large demonstrates yet another pattern: the
performance is low for almost all parameter combinations, but then
suddenly jumps to 0.74 at the last layer. (The results on
layer 23, not shown in the table, are very similar to those from layer
22.) Additionally, T5 seems to encode semantics in a
higher-dimensional subspace, with projecting on 128 dimensions being
the minimum and 1024 still working well.

\begin{table}[t]
\centering
\resizebox{0.48\textwidth}{!}{
\begin{tabular}{rrrrrrrrrrrrr}

\rowcolor[HTML]{FFFFFF} 
\multicolumn{13}{c}{\cellcolor[HTML]{FFFFFF}\textbf{GPT-2 large}} \\
\rowcolor[HTML]{FFFFFF} 
\multicolumn{1}{l}{\cellcolor[HTML]{FFFFFF}} &
  \textbf{3} &
  \textbf{6} &
  \textbf{9} &
  \textbf{12} &
  \textbf{15} &
  \textbf{18} &
  \textbf{21} &
  \textbf{24} &
  \textbf{27} &
  \textbf{30} &
  \textbf{33} &
  \textbf{36} \\ \cline{2-13} 
\multicolumn{1}{r|}{\cellcolor[HTML]{FFFFFF}\textbf{2}} &
  \cellcolor[HTML]{FCEFEE}0.23 &
  \cellcolor[HTML]{FBEAE8}0.25 &
  \cellcolor[HTML]{FDF5F4}0.21 &
  \cellcolor[HTML]{FCEBEA}0.24 &
  \cellcolor[HTML]{F8D8D5}0.3 &
  \cellcolor[HTML]{F7D4D1}0.31 &
  \cellcolor[HTML]{FCECEA}0.24 &
  \cellcolor[HTML]{FBE8E7}0.25 &
  \cellcolor[HTML]{FFFDFD}0.19 &
  \cellcolor[HTML]{FFFFFF}0.18 &
  \cellcolor[HTML]{FDF3F2}0.22 &
  \cellcolor[HTML]{F3BFBA}0.38 \\
\multicolumn{1}{r|}{\cellcolor[HTML]{FFFFFF}\textbf{4}} &
  \cellcolor[HTML]{F2BBB7}0.39 &
  \cellcolor[HTML]{F4C2BE}0.37 &
  \cellcolor[HTML]{F4C2BE}0.36 &
  \cellcolor[HTML]{F4C3BF}0.36 &
  \cellcolor[HTML]{F3C0BC}0.37 &
  \cellcolor[HTML]{F6CECA}0.33 &
  \cellcolor[HTML]{F9DCDA}0.29 &
  \cellcolor[HTML]{FBE6E4}0.26 &
  \cellcolor[HTML]{FDF0EF}0.23 &
  \cellcolor[HTML]{FCEEED}0.23 &
  \cellcolor[HTML]{FAE3E1}0.27 &
  \cellcolor[HTML]{EC9992}0.49 \\
\multicolumn{1}{r|}{\cellcolor[HTML]{FFFFFF}\textbf{8}} &
  \cellcolor[HTML]{EEA49E}0.45 &
  \cellcolor[HTML]{F2B6B1}0.4 &
  \cellcolor[HTML]{F3BEB9}0.38 &
  \cellcolor[HTML]{F1B5AF}0.41 &
  \cellcolor[HTML]{F1B5B0}0.4 &
  \cellcolor[HTML]{F9DCDA}0.29 &
  \cellcolor[HTML]{F8DAD8}0.29 &
  \cellcolor[HTML]{FBE5E4}0.26 &
  \cellcolor[HTML]{FAE3E1}0.27 &
  \cellcolor[HTML]{FAE3E1}0.27 &
  \cellcolor[HTML]{F8D9D6}0.3 &
  \cellcolor[HTML]{E8837B}0.55 \\
\multicolumn{1}{r|}{\cellcolor[HTML]{FFFFFF}\textbf{16}} &
  \cellcolor[HTML]{EEA29C}0.46 &
  \cellcolor[HTML]{F2B7B3}0.4 &
  \cellcolor[HTML]{F3BCB7}0.38 &
  \cellcolor[HTML]{F1B4AF}0.41 &
  \cellcolor[HTML]{F3BEBA}0.38 &
  \cellcolor[HTML]{FAE0DE}0.27 &
  \cellcolor[HTML]{F9DBD9}0.29 &
  \cellcolor[HTML]{F9E0DE}0.28 &
  \cellcolor[HTML]{F9E0DE}0.28 &
  \cellcolor[HTML]{F9DDDB}0.28 &
  \cellcolor[HTML]{F7D2CE}0.32 &
  \cellcolor[HTML]{E67C73}\textbf{0.57} \\
\multicolumn{1}{r|}{\cellcolor[HTML]{FFFFFF}\textbf{32}} &
  \cellcolor[HTML]{EFA9A3}0.44 &
  \cellcolor[HTML]{F3BBB7}0.38 &
  \cellcolor[HTML]{F3C1BC}0.37 &
  \cellcolor[HTML]{F3BDB8}0.38 &
  \cellcolor[HTML]{FBE7E5}0.26 &
  \cellcolor[HTML]{F9DBD9}0.29 &
  \cellcolor[HTML]{F8D9D6}0.3 &
  \cellcolor[HTML]{F9DEDC}0.28 &
  \cellcolor[HTML]{F8DAD8}0.29 &
  \cellcolor[HTML]{F8DAD7}0.29 &
  \cellcolor[HTML]{F6CFCC}0.33 &
  \cellcolor[HTML]{E78178}\textbf{0.56} \\
\multicolumn{1}{r|}{\cellcolor[HTML]{FFFFFF}\textbf{64}} &
  \cellcolor[HTML]{F0AFA9}0.42 &
  \cellcolor[HTML]{F2B9B4}0.39 &
  \cellcolor[HTML]{F4C3BF}0.36 &
  \cellcolor[HTML]{FCEFEE}0.23 &
  \cellcolor[HTML]{FAE5E3}0.26 &
  \cellcolor[HTML]{F8D9D6}0.3 &
  \cellcolor[HTML]{F8D6D4}0.3 &
  \cellcolor[HTML]{F9DDDB}0.28 &
  \cellcolor[HTML]{F9DBD9}0.29 &
  \cellcolor[HTML]{F8D7D5}0.3 &
  \cellcolor[HTML]{F6CDC9}0.33 &
  \cellcolor[HTML]{F6CECA}0.33 \\
\multicolumn{1}{r|}{\cellcolor[HTML]{FFFFFF}\textbf{128}} &
  \cellcolor[HTML]{F0AEA8}0.43 &
  \cellcolor[HTML]{F5C6C2}0.35 &
  \cellcolor[HTML]{FEF8F7}0.2 &
  \cellcolor[HTML]{FCEEED}0.23 &
  \cellcolor[HTML]{FAE4E2}0.26 &
  \cellcolor[HTML]{F8D9D6}0.3 &
  \cellcolor[HTML]{F7D5D3}0.31 &
  \cellcolor[HTML]{F9DBD9}0.29 &
  \cellcolor[HTML]{F8DAD7}0.29 &
  \cellcolor[HTML]{F8D8D5}0.3 &
  \cellcolor[HTML]{F6CCC8}0.34 &
  \cellcolor[HTML]{F6D0CD}0.32 \\
\multicolumn{1}{r|}{\cellcolor[HTML]{FFFFFF}\textbf{256}} &
  \cellcolor[HTML]{F8D9D6}0.3 &
  \cellcolor[HTML]{FEF5F5}0.21 &
  \cellcolor[HTML]{FFFBFB}0.19 &
  \cellcolor[HTML]{FCEFEE}0.23 &
  \cellcolor[HTML]{FAE3E1}0.27 &
  \cellcolor[HTML]{F8D8D5}0.3 &
  \cellcolor[HTML]{F7D4D2}0.31 &
  \cellcolor[HTML]{F9DCD9}0.29 &
  \cellcolor[HTML]{F8D9D7}0.3 &
  \cellcolor[HTML]{F8D7D5}0.3 &
  \cellcolor[HTML]{F6CBC8}0.34 &
  \cellcolor[HTML]{F7D0CD}0.32 \\
\multicolumn{1}{r|}{\cellcolor[HTML]{FFFFFF}\textbf{512}} &
  \cellcolor[HTML]{F8D9D7}0.3 &
  \cellcolor[HTML]{FEF6F5}0.21 &
  \cellcolor[HTML]{FFFBFB}0.19 &
  \cellcolor[HTML]{FCEFEE}0.23 &
  \cellcolor[HTML]{FAE3E1}0.27 &
  \cellcolor[HTML]{F8D7D4}0.3 &
  \cellcolor[HTML]{F7D3D0}0.31 &
  \cellcolor[HTML]{F9DCD9}0.29 &
  \cellcolor[HTML]{F8D9D7}0.3 &
  \cellcolor[HTML]{F8D7D4}0.3 &
  \cellcolor[HTML]{F6CBC8}0.34 &
  \cellcolor[HTML]{F7D1CE}0.32 \\
\multicolumn{1}{r|}{\cellcolor[HTML]{FFFFFF}\textbf{1280}} &
  \cellcolor[HTML]{F8DAD7}0.29 &
  \cellcolor[HTML]{FEF6F6}0.21 &
  \cellcolor[HTML]{FFFBFB}0.2 &
  \cellcolor[HTML]{FCEFEE}0.23 &
  \cellcolor[HTML]{FAE3E1}0.27 &
  \cellcolor[HTML]{F8D7D4}0.3 &
  \cellcolor[HTML]{F7D4D1}0.31 &
  \cellcolor[HTML]{F9DBD9}0.29 &
  \cellcolor[HTML]{F8D9D6}0.3 &
  \cellcolor[HTML]{F8D7D4}0.3 &
  \cellcolor[HTML]{F5CBC7}0.34 &
  \cellcolor[HTML]{F7D2CF}0.32 \\
\multicolumn{1}{r|}{\cellcolor[HTML]{FFFFFF}\textbf{1280$'$}} &
  \cellcolor[HTML]{F9DEDC}0.28 &
  \cellcolor[HTML]{FEF8F7}0.21 &
  \cellcolor[HTML]{FFFCFB}0.19 &
  \cellcolor[HTML]{FDF2F1}0.22 &
  \cellcolor[HTML]{FAE3E1}0.27 &
  \cellcolor[HTML]{F8D7D4}0.3 &
  \cellcolor[HTML]{F7D3D0}0.31 &
  \cellcolor[HTML]{F8DAD7}0.29 &
  \cellcolor[HTML]{F8D6D3}0.31 &
  \cellcolor[HTML]{F7D2CF}0.32 &
  \cellcolor[HTML]{F5C8C4}0.35 &
  \cellcolor[HTML]{F7D4D1}0.31 \\

\rowcolor[HTML]{FFFFFF} 
\multicolumn{13}{c}{\cellcolor[HTML]{FFFFFF}\textbf{Llama 7B}} \\
\rowcolor[HTML]{FFFFFF} 
\multicolumn{1}{l}{\cellcolor[HTML]{FFFFFF}} &
  \textbf{1} &
  \textbf{2} &
  \textbf{5} &
  \textbf{8} &
  \textbf{11} &
  \textbf{14} &
  \textbf{17} &
  \textbf{20} &
  \textbf{23} &
  \textbf{26} &
  \textbf{29} &
  \textbf{32} \\ \cline{2-13} 
\multicolumn{1}{r|}{\cellcolor[HTML]{FFFFFF}\textbf{2}} &
  \cellcolor[HTML]{FFFCFC}0.11 &
  \cellcolor[HTML]{FFFFFF}0.09 &
  \cellcolor[HTML]{F9E0DD}0.21 &
  \cellcolor[HTML]{FBEAE9}0.17 &
  \cellcolor[HTML]{F8D8D5}0.24 &
  \cellcolor[HTML]{F2B7B2}0.37 &
  \cellcolor[HTML]{F2B9B5}0.36 &
  \cellcolor[HTML]{F2BAB5}0.35 &
  \cellcolor[HTML]{F5C9C6}0.3 &
  \cellcolor[HTML]{F8DAD7}0.24 &
  \cellcolor[HTML]{F8DAD7}0.23 &
  \cellcolor[HTML]{EFA7A1}0.43 \\
\multicolumn{1}{r|}{\cellcolor[HTML]{FFFFFF}\textbf{4}} &
  \cellcolor[HTML]{FEFAFA}0.11 &
  \cellcolor[HTML]{FEF9F8}0.12 &
  \cellcolor[HTML]{F3BFBA}0.34 &
  \cellcolor[HTML]{F8D8D5}0.24 &
  \cellcolor[HTML]{EB948D}0.5 &
  \cellcolor[HTML]{EB968F}0.49 &
  \cellcolor[HTML]{ED9F99}0.45 &
  \cellcolor[HTML]{EEA39C}0.44 &
  \cellcolor[HTML]{F3C0BC}0.33 &
  \cellcolor[HTML]{F7D1CD}0.27 &
  \cellcolor[HTML]{F7D3D0}0.26 &
  \cellcolor[HTML]{E98C84}0.53 \\
\multicolumn{1}{r|}{\cellcolor[HTML]{FFFFFF}\textbf{8}} &
  \cellcolor[HTML]{FFFAFA}0.11 &
  \cellcolor[HTML]{F8D9D7}0.24 &
  \cellcolor[HTML]{EFA9A4}0.42 &
  \cellcolor[HTML]{EEA39C}0.44 &
  \cellcolor[HTML]{E8837B}0.56 &
  \cellcolor[HTML]{E8857D}0.55 &
  \cellcolor[HTML]{EA8F87}0.52 &
  \cellcolor[HTML]{F0ADA7}0.4 &
  \cellcolor[HTML]{F6CDCA}0.28 &
  \cellcolor[HTML]{F5CAC6}0.3 &
  \cellcolor[HTML]{F5C8C4}0.3 &
  \cellcolor[HTML]{E78077}\textbf{0.57} \\
\multicolumn{1}{r|}{\cellcolor[HTML]{FFFFFF}\textbf{16}} &
  \cellcolor[HTML]{FFFAFA}0.11 &
  \cellcolor[HTML]{FCECEA}0.17 &
  \cellcolor[HTML]{EC9C95}0.47 &
  \cellcolor[HTML]{E8837B}0.56 &
  \cellcolor[HTML]{E67C73}\textbf{0.58} &
  \cellcolor[HTML]{E8837B}0.56 &
  \cellcolor[HTML]{EB928A}0.5 &
  \cellcolor[HTML]{F5C8C4}0.3 &
  \cellcolor[HTML]{F6CCC8}0.29 &
  \cellcolor[HTML]{F4C3BF}0.32 &
  \cellcolor[HTML]{F3C1BC}0.33 &
  \cellcolor[HTML]{E98981}0.54 \\
\multicolumn{1}{r|}{\cellcolor[HTML]{FFFFFF}\textbf{32}} &
  \cellcolor[HTML]{FEFAF9}0.12 &
  \cellcolor[HTML]{F5C8C4}0.3 &
  \cellcolor[HTML]{EB928A}0.51 &
  \cellcolor[HTML]{E8847B}0.56 &
  \cellcolor[HTML]{E78077}\textbf{0.57} &
  \cellcolor[HTML]{EA8C84}0.53 &
  \cellcolor[HTML]{F5C8C4}0.3 &
  \cellcolor[HTML]{F5C8C4}0.3 &
  \cellcolor[HTML]{F5C9C6}0.3 &
  \cellcolor[HTML]{F4C2BE}0.32 &
  \cellcolor[HTML]{F3BFBA}0.34 &
  \cellcolor[HTML]{F2BAB5}0.35 \\
\multicolumn{1}{r|}{\cellcolor[HTML]{FFFFFF}\textbf{64}} &
  \cellcolor[HTML]{FEFAF9}0.12 &
  \cellcolor[HTML]{F0B1AB}0.39 &
  \cellcolor[HTML]{EA8F88}0.51 &
  \cellcolor[HTML]{E9877F}0.54 &
  \cellcolor[HTML]{E98880}0.54 &
  \cellcolor[HTML]{F4C5C1}0.31 &
  \cellcolor[HTML]{F5C9C5}0.3 &
  \cellcolor[HTML]{F5CAC6}0.3 &
  \cellcolor[HTML]{F5C8C4}0.3 &
  \cellcolor[HTML]{F4C2BE}0.32 &
  \cellcolor[HTML]{F3BEB9}0.34 &
  \cellcolor[HTML]{F3BCB7}0.35 \\
\multicolumn{1}{r|}{\cellcolor[HTML]{FFFFFF}\textbf{128}} &
  \cellcolor[HTML]{FEF9F8}0.12 &
  \cellcolor[HTML]{EFABA6}0.41 &
  \cellcolor[HTML]{EA9089}0.51 &
  \cellcolor[HTML]{EA8F87}0.52 &
  \cellcolor[HTML]{EB958D}0.49 &
  \cellcolor[HTML]{F5C8C5}0.3 &
  \cellcolor[HTML]{F5CAC7}0.29 &
  \cellcolor[HTML]{F5C9C6}0.3 &
  \cellcolor[HTML]{F5C8C4}0.3 &
  \cellcolor[HTML]{F4C1BD}0.33 &
  \cellcolor[HTML]{F3BDB9}0.34 &
  \cellcolor[HTML]{F3BEB9}0.34 \\
\multicolumn{1}{r|}{\cellcolor[HTML]{FFFFFF}\textbf{256}} &
  \cellcolor[HTML]{FEF7F7}0.12 &
  \cellcolor[HTML]{EFA9A3}0.42 &
  \cellcolor[HTML]{EB938B}0.5 &
  \cellcolor[HTML]{EC9790}0.49 &
  \cellcolor[HTML]{F4C6C2}0.31 &
  \cellcolor[HTML]{F5CAC6}0.29 &
  \cellcolor[HTML]{F5CBC7}0.29 &
  \cellcolor[HTML]{F5C9C6}0.3 &
  \cellcolor[HTML]{F5C8C4}0.3 &
  \cellcolor[HTML]{F4C1BD}0.33 &
  \cellcolor[HTML]{F3BDB8}0.34 &
  \cellcolor[HTML]{F3BFBA}0.34 \\
\multicolumn{1}{r|}{\cellcolor[HTML]{FFFFFF}\textbf{512}} &
  \cellcolor[HTML]{FEF5F5}0.13 &
  \cellcolor[HTML]{EEA59F}0.43 &
  \cellcolor[HTML]{EB978F}0.49 &
  \cellcolor[HTML]{EEA49E}0.44 &
  \cellcolor[HTML]{F5C8C4}0.3 &
  \cellcolor[HTML]{F6CBC8}0.29 &
  \cellcolor[HTML]{F6CBC8}0.29 &
  \cellcolor[HTML]{F5C9C6}0.3 &
  \cellcolor[HTML]{F5C7C3}0.3 &
  \cellcolor[HTML]{F4C1BC}0.33 &
  \cellcolor[HTML]{F3BDB8}0.34 &
  \cellcolor[HTML]{F3BFBB}0.33 \\
\multicolumn{1}{r|}{\cellcolor[HTML]{FFFFFF}\textbf{4096}} &
  \cellcolor[HTML]{FBE8E6}0.18 &
  \cellcolor[HTML]{EEA39D}0.44 &
  \cellcolor[HTML]{F0AEA8}0.4 &
  \cellcolor[HTML]{FBE5E4}0.19 &
  \cellcolor[HTML]{F5CBC7}0.29 &
  \cellcolor[HTML]{F6CCC9}0.29 &
  \cellcolor[HTML]{F6CCC9}0.29 &
  \cellcolor[HTML]{F5CAC6}0.29 &
  \cellcolor[HTML]{F5C8C4}0.3 &
  \cellcolor[HTML]{F4C1BD}0.33 &
  \cellcolor[HTML]{F3BDB8}0.34 &
  \cellcolor[HTML]{F3C1BC}0.33 \\
\multicolumn{1}{r|}{\cellcolor[HTML]{FFFFFF}\textbf{4096$'$}} &
  \cellcolor[HTML]{FAE2E0}0.2 &
  \cellcolor[HTML]{FDF0EF}0.15 &
  \cellcolor[HTML]{FCEEED}0.16 &
  \cellcolor[HTML]{FBE7E5}0.19 &
  \cellcolor[HTML]{F3BFBB}0.34 &
  \cellcolor[HTML]{F2BBB6}0.35 &
  \cellcolor[HTML]{F2BAB5}0.35 &
  \cellcolor[HTML]{F2B9B4}0.36 &
  \cellcolor[HTML]{F2B8B3}0.36 &
  \cellcolor[HTML]{F0ACA6}0.41 &
  \cellcolor[HTML]{EEA59F}0.43 &
  \cellcolor[HTML]{F2BAB5}0.36 \\
  
\rowcolor[HTML]{FFFFFF} 
\multicolumn{13}{c}{\cellcolor[HTML]{FFFFFF}\textbf{OPT 30B}} \\
\rowcolor[HTML]{FFFFFF} 
\multicolumn{1}{l}{\cellcolor[HTML]{FFFFFF}} &
  \textbf{4} &
  \textbf{8} &
  \textbf{12} &
  \textbf{16} &
  \textbf{20} &
  \textbf{24} &
  \textbf{28} &
  \textbf{32} &
  \textbf{36} &
  \textbf{40} &
  \textbf{44} &
  \textbf{48} \\ \cline{2-13} 
\multicolumn{1}{r|}{\cellcolor[HTML]{FFFFFF}\textbf{2}} &
  \cellcolor[HTML]{FFFFFF}0.08 &
  \cellcolor[HTML]{FDF1F0}0.13 &
  \cellcolor[HTML]{FDF3F2}0.13 &
  \cellcolor[HTML]{FDF5F4}0.12 &
  \cellcolor[HTML]{FBE9E7}0.16 &
  \cellcolor[HTML]{F0AFA9}0.38 &
  \cellcolor[HTML]{F0ACA7}0.38 &
  \cellcolor[HTML]{F3BCB7}0.33 &
  \cellcolor[HTML]{F4C2BE}0.3 &
  \cellcolor[HTML]{F8D7D4}0.23 &
  \cellcolor[HTML]{F9DBD9}0.21 &
  \cellcolor[HTML]{EEA49E}0.41 \\
\multicolumn{1}{r|}{\cellcolor[HTML]{FFFFFF}\textbf{4}} &
  \cellcolor[HTML]{FCF0EF}0.14 &
  \cellcolor[HTML]{FDF4F3}0.12 &
  \cellcolor[HTML]{FCECEB}0.15 &
  \cellcolor[HTML]{FCECEB}0.15 &
  \cellcolor[HTML]{FBE7E5}0.17 &
  \cellcolor[HTML]{ED9C95}0.44 &
  \cellcolor[HTML]{E98B83}0.5 &
  \cellcolor[HTML]{EC9790}0.46 &
  \cellcolor[HTML]{F4C4C0}0.3 &
  \cellcolor[HTML]{F5C7C3}0.29 &
  \cellcolor[HTML]{F5CBC7}0.27 &
  \cellcolor[HTML]{E8847C}0.53 \\
\multicolumn{1}{r|}{\cellcolor[HTML]{FFFFFF}\textbf{8}} &
  \cellcolor[HTML]{FCEDEC}0.15 &
  \cellcolor[HTML]{FCEDEC}0.15 &
  \cellcolor[HTML]{FBE9E7}0.16 &
  \cellcolor[HTML]{FBE8E6}0.17 &
  \cellcolor[HTML]{F9E0DD}0.2 &
  \cellcolor[HTML]{E98880}0.52 &
  \cellcolor[HTML]{E98A82}0.51 &
  \cellcolor[HTML]{EC9C95}0.44 &
  \cellcolor[HTML]{F2BBB6}0.33 &
  \cellcolor[HTML]{F3BEB9}0.32 &
  \cellcolor[HTML]{F2BBB6}0.33 &
  \cellcolor[HTML]{E67C73}\textbf{0.56} \\
\multicolumn{1}{r|}{\cellcolor[HTML]{FFFFFF}\textbf{16}} &
  \cellcolor[HTML]{FBE5E4}0.18 &
  \cellcolor[HTML]{FBEAE8}0.16 &
  \cellcolor[HTML]{FBE6E4}0.17 &
  \cellcolor[HTML]{FBE7E5}0.17 &
  \cellcolor[HTML]{FAE2E0}0.19 &
  \cellcolor[HTML]{E98981}0.51 &
  \cellcolor[HTML]{EA8D85}0.5 &
  \cellcolor[HTML]{F2B7B2}0.34 &
  \cellcolor[HTML]{F2B8B3}0.34 &
  \cellcolor[HTML]{F2BAB5}0.33 &
  \cellcolor[HTML]{F1B4AF}0.35 &
  \cellcolor[HTML]{E8857C}0.53 \\
\multicolumn{1}{r|}{\cellcolor[HTML]{FFFFFF}\textbf{32}} &
  \cellcolor[HTML]{FAE2DF}0.19 &
  \cellcolor[HTML]{F9E0DE}0.19 &
  \cellcolor[HTML]{FAE1DF}0.19 &
  \cellcolor[HTML]{FBE7E5}0.17 &
  \cellcolor[HTML]{FAE1DF}0.19 &
  \cellcolor[HTML]{EA8E86}0.5 &
  \cellcolor[HTML]{F1B4AF}0.36 &
  \cellcolor[HTML]{F2BAB5}0.33 &
  \cellcolor[HTML]{F2B6B1}0.35 &
  \cellcolor[HTML]{F2B9B4}0.34 &
  \cellcolor[HTML]{F0B0AB}0.37 &
  \cellcolor[HTML]{F1B4AF}0.35 \\
\multicolumn{1}{r|}{\cellcolor[HTML]{FFFFFF}\textbf{64}} &
  \cellcolor[HTML]{F7D4D1}0.24 &
  \cellcolor[HTML]{F9DDDB}0.21 &
  \cellcolor[HTML]{F9DCDA}0.21 &
  \cellcolor[HTML]{FAE1DF}0.19 &
  \cellcolor[HTML]{F9DDDA}0.21 &
  \cellcolor[HTML]{EC9A93}0.45 &
  \cellcolor[HTML]{F1B4AF}0.36 &
  \cellcolor[HTML]{F2B6B1}0.35 &
  \cellcolor[HTML]{F2B7B2}0.35 &
  \cellcolor[HTML]{F1B4AF}0.36 &
  \cellcolor[HTML]{F1B1AC}0.37 &
  \cellcolor[HTML]{F1B5B0}0.35 \\
\multicolumn{1}{r|}{\cellcolor[HTML]{FFFFFF}\textbf{128}} &
  \cellcolor[HTML]{E9877F}0.52 &
  \cellcolor[HTML]{F7D1CE}0.25 &
  \cellcolor[HTML]{F8D6D3}0.23 &
  \cellcolor[HTML]{F9DCDA}0.21 &
  \cellcolor[HTML]{EFA8A2}0.4 &
  \cellcolor[HTML]{F0ACA7}0.38 &
  \cellcolor[HTML]{F1B5B0}0.35 &
  \cellcolor[HTML]{F2B9B4}0.34 &
  \cellcolor[HTML]{F1B4AF}0.35 &
  \cellcolor[HTML]{F1B5B0}0.35 &
  \cellcolor[HTML]{F0B1AB}0.37 &
  \cellcolor[HTML]{F2B8B3}0.34 \\
\multicolumn{1}{r|}{\cellcolor[HTML]{FFFFFF}\textbf{256}} &
  \cellcolor[HTML]{E98C84}0.5 &
  \cellcolor[HTML]{EB958E}0.47 &
  \cellcolor[HTML]{EEA29C}0.42 &
  \cellcolor[HTML]{F0ACA6}0.38 &
  \cellcolor[HTML]{F2B8B3}0.34 &
  \cellcolor[HTML]{F3BFBA}0.32 &
  \cellcolor[HTML]{F2B6B1}0.35 &
  \cellcolor[HTML]{F2B9B4}0.34 &
  \cellcolor[HTML]{F1B4AF}0.35 &
  \cellcolor[HTML]{F1B5B0}0.35 &
  \cellcolor[HTML]{F0AEA9}0.38 &
  \cellcolor[HTML]{F2B9B5}0.34 \\
\multicolumn{1}{r|}{\cellcolor[HTML]{FFFFFF}\textbf{512}} &
  \cellcolor[HTML]{EB938B}0.48 &
  \cellcolor[HTML]{EC9B94}0.45 &
  \cellcolor[HTML]{F1B5B0}0.35 &
  \cellcolor[HTML]{F3BCB7}0.33 &
  \cellcolor[HTML]{F4C1BD}0.31 &
  \cellcolor[HTML]{F6CCC8}0.27 &
  \cellcolor[HTML]{F2B7B2}0.34 &
  \cellcolor[HTML]{F2B9B4}0.34 &
  \cellcolor[HTML]{F1B5B0}0.35 &
  \cellcolor[HTML]{F1B5B0}0.35 &
  \cellcolor[HTML]{F0ADA8}0.38 &
  \cellcolor[HTML]{F2B9B4}0.34 \\
\multicolumn{1}{r|}{\cellcolor[HTML]{FFFFFF}\textbf{7168}} &
  \cellcolor[HTML]{F4C4C0}0.3 &
  \cellcolor[HTML]{F5C7C3}0.29 &
  \cellcolor[HTML]{F5C9C5}0.28 &
  \cellcolor[HTML]{F6CFCB}0.26 &
  \cellcolor[HTML]{F8D7D4}0.23 &
  \cellcolor[HTML]{F6CFCC}0.26 &
  \cellcolor[HTML]{F2B9B4}0.34 &
  \cellcolor[HTML]{F2BAB5}0.33 &
  \cellcolor[HTML]{F1B6B1}0.35 &
  \cellcolor[HTML]{F1B4AF}0.35 &
  \cellcolor[HTML]{F0AEA8}0.38 &
  \cellcolor[HTML]{F2BBB6}0.33 \\
\multicolumn{1}{r|}{\cellcolor[HTML]{FFFFFF}\textbf{7168$'$}} &
  \cellcolor[HTML]{FAE1DF}0.19 &
  \cellcolor[HTML]{FAE0DE}0.19 &
  \cellcolor[HTML]{F9DEDB}0.2 &
  \cellcolor[HTML]{FAE0DE}0.19 &
  \cellcolor[HTML]{F9DFDD}0.2 &
  \cellcolor[HTML]{F5CAC6}0.28 &
  \cellcolor[HTML]{EEA6A0}0.41 &
  \cellcolor[HTML]{EEA19B}0.42 &
  \cellcolor[HTML]{ED9E97}0.44 &
  \cellcolor[HTML]{EC9891}0.46 &
  \cellcolor[HTML]{EA8F87}0.49 &
  \cellcolor[HTML]{EFAAA4}0.39
\end{tabular}
}
\caption{Spearman correlations of sentence-similarity scores derived via projection from 
averaged-token representations by model, layer (columns), and subspace dimensionality (rows)
with the STS benchmark scores. $'$ stands for using vanilla 
representations without projection. The results are averaged over ten runs.}
\label{tab:sts-gpt-llama}
\end{table}

\begin{figure*}[ht]
    \centering
    \includegraphics[width=\linewidth]{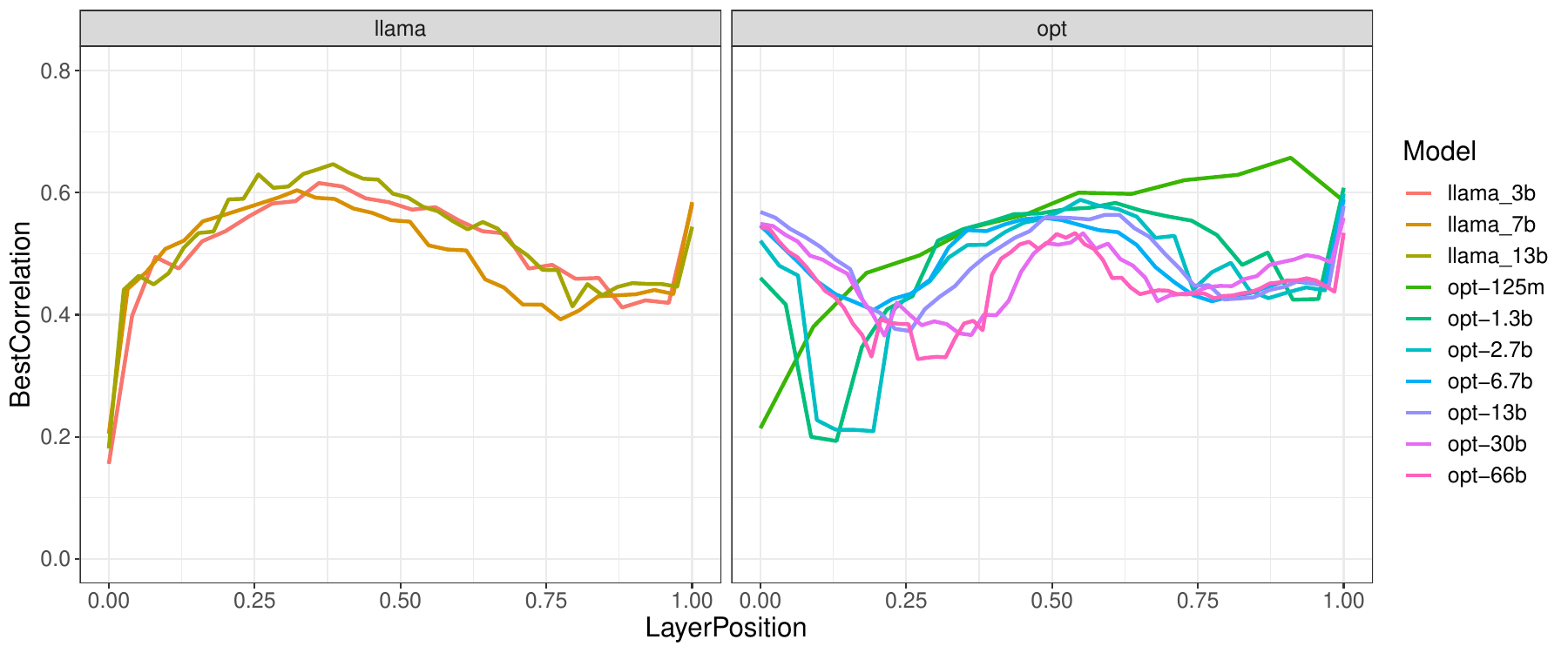}
    \caption{Performance of Llama and OPT on the STS probing task by layer and model size.}
    \label{fig:sts-perf-opt-t5-by-size}
\end{figure*}

The results from causal models, shown in
Table~\ref{tab:sts-gpt-llama}, demonstrate a different consistent  pattern. 
The best performance is worse, \(r = \) 0.56--0.58, and it
is usually achieved in the last layer, similarly to T5, but the
optimal dimensionality of the projections is much lower (between~8
and~32), despite the models' higher embedding dimensionality. Also,
there is a clear cyclic development in performance across
layers. E.g., with a dimensionality of 16, GPT-2 large first goes high
(0.46), then low ($\approx$~0.28), then high again (0.57). Llama 7B
shows 1.5 cycles and OPT 30B two full cycles
(cf.\ also Figure~\ref{fig:sts-perf-opt-t5-by-size}).

The relatively lower informativeness of causal models' representations compared to those
provided by MLMs seems to support the argument that they are less suited for representation
learning \citep{clark2020electra,reimers2022embeddings}. However, the fact that they are most 
informative in the last layer goes against the  previous interpretation that the last layer of
GPT-2 undergoes a representation collapse \citep{ethayarajh-2019-contextual} and rather supports 
the argument that the extreme anisotropy of the last layer of GPT-2 is an artefact of inadequate 
similarity modelling \citep{timkey-van-schijndel-2021-bark}.\footnote{Figures~\ref{fig:perf-by-layer} and \ref{fig:perf-by-dimension} in the Appendix provide a visualisation of the distribution of the performance 
by the normalised layer position and projection-space dimensionality across models.}

\begin{figure*}[t]
    \centering
    \includegraphics[width=\linewidth]{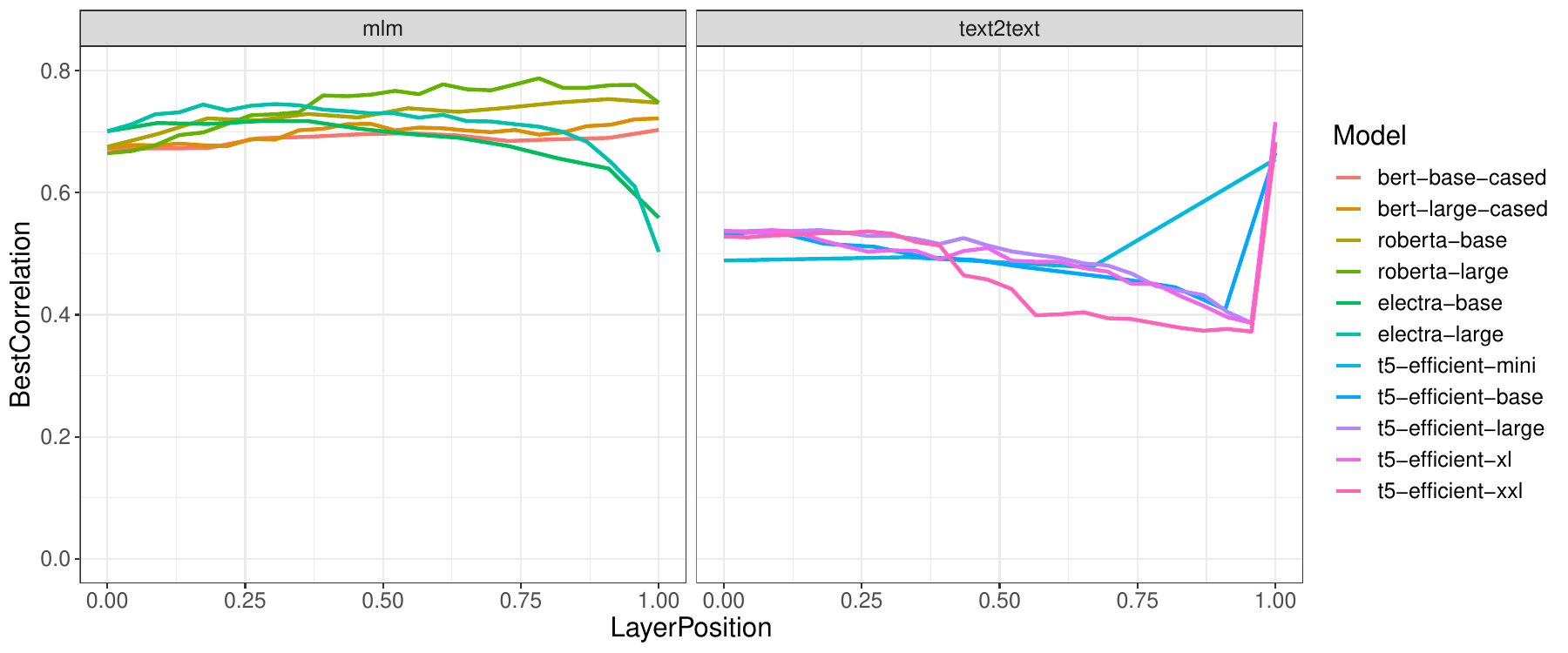}
    \caption{Performance of encoder-only and text-to-text  MLMs on the  STS probing task by layer and model size.}
    \label{fig:sts-perf-mlms-by-size}
\end{figure*}

\paragraph{Results across model sizes} 
We now study the connection between the semantic content of models' 
representations, as measured by our structural probe, and their size. Classic MLMs, 
such as BERT and RoBERTa, are only available in a few sizes, not counting 
various smaller distilled versions, such as TinyBERT \citep{jiao-etal-2020-tinybert}.
Later models (both text-to-text and causal) were published in a larger size range. 
Aggregated results are presented visually in Figure~\ref{fig:sts-perf-mlms-by-size}
(cf. Section~\ref{ssec:presentation}).
The left pane shows the performance of encoder-only
MLMs across two model sizes each. We see that while bigger
models perform better, the distribution of the semantic information
across layers is very similar across model sizes. This finding is further strengthened 
by the analysis of the performance of the T5-efficient, Llama, and OPT models shown in the 
right-hand pane of Figure\ \ref{fig:sts-perf-mlms-by-size}. All T5-efficient models attain the best 
performance in the very last layer and show some loss of informativity in middle layers. 

The three Llama models, shown in the left pane of 
Figure~\ref{fig:sts-perf-opt-t5-by-size}, follow the pattern from Table~\ref{tab:sts-gpt-llama}:
the performance of the lower layers is almost zero, while middle layers attain maximum performance, 
which then decreases and goes up again at the end. The behaviour of the OPT models is even more 
complicated: nearly all of them demonstrate the \enquote{double dip} pattern, where as the
layer number increases the performance first goes down, then up, then
down again, and finally reaches the peak in the last layer.  This
oscillating pattern in the performance of causal LMs does not fully align 
with the the conclusions reached by \citet{geva-etal-2021-transformer}, who claim
that there exists a progression of lexical, syntactic, and semantic features 
as information flows through decoder-only models language models.

OPT, interestingly, is also the only model class where we see a
substantial effect of model size: the smallest model, OPT 125m, shows
a steady increase in performance with a slight drop at the very
end. It outperforms all larger models and nearly reaches the results
of T5. This seems to suggest that extremely small causal LMs have
non-trivial representation-learning capacities.

\subsubsection{Natural Language Inference}\label{ssec:results-nli}

In this section, we check if the distribution of semantic information across
model layers we identified in the context of the STS task can also be observed in the context of 
NLI. We further check if
the patterns are dependent on the dataset and on the exact operationalization: 
we contrast triplet-focused 
probing, which is structurally close to our 
STS operationalization, with cosine-similarity-based probing, which operates on the level of
sentence pairs and permits us to cover more data.

\begin{figure*}[t]
    \centering
    \includegraphics[width=\linewidth]{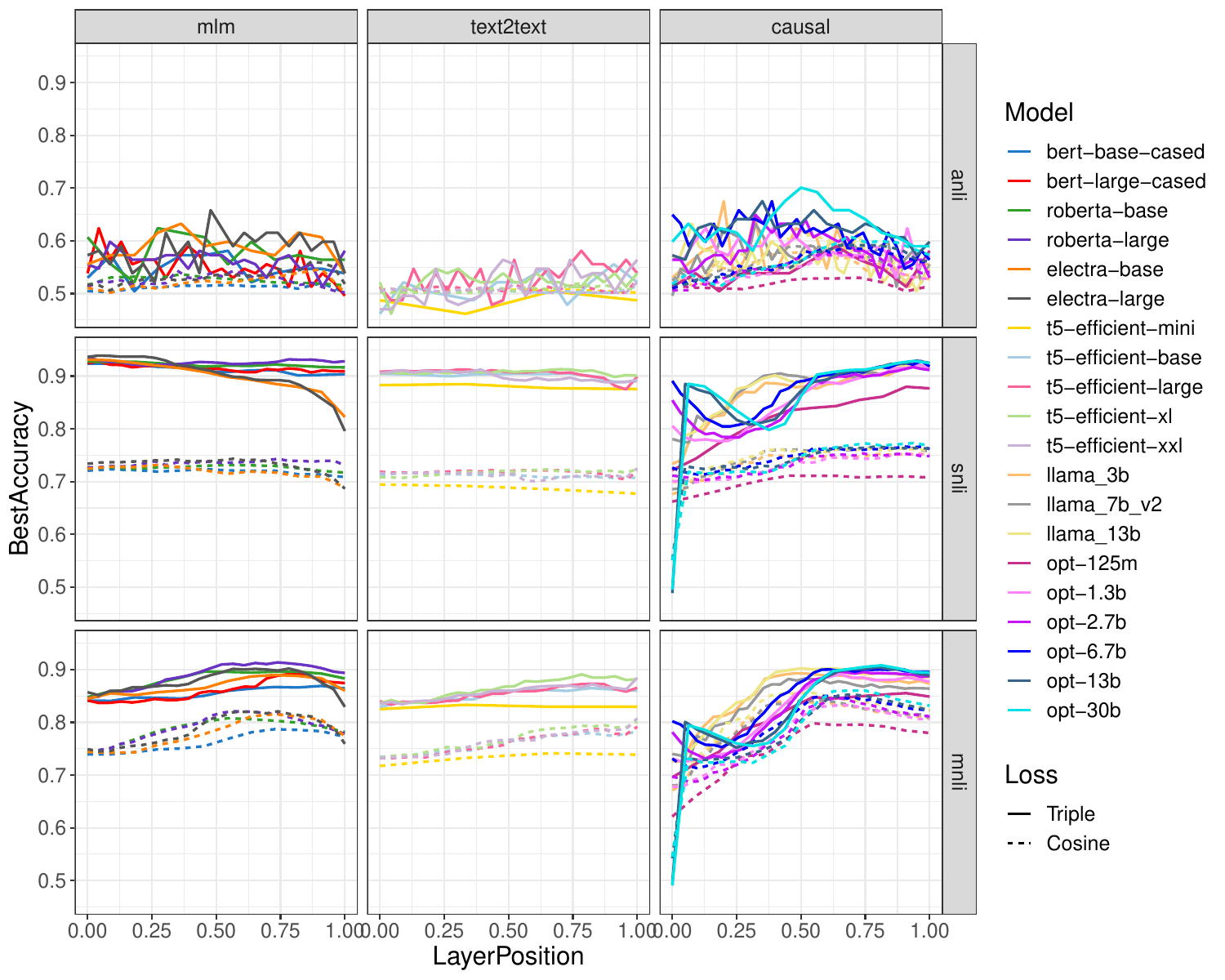}
    \caption{Performance of models on NLI across datasets, 
    model types and sizes, and model layers.}\label{fig:nli-results}
\end{figure*}

\begin{figure*}[t]
    \centering
    \includegraphics[width=\linewidth]{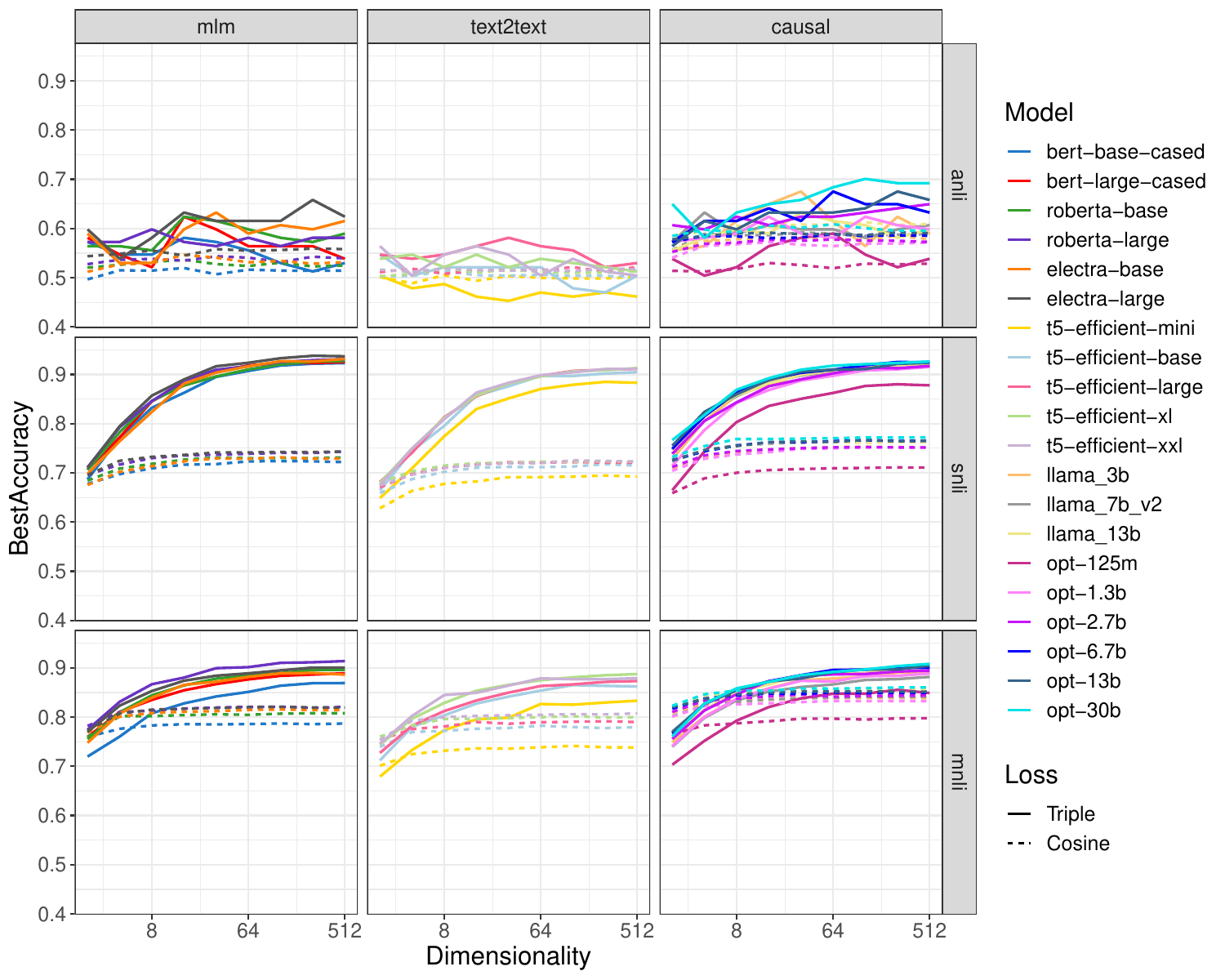}
    \caption{Performance of models on NLI across datasets,
    model types and sizes, and projection-space dimensions.}\label{fig:nli-results-dim}
\end{figure*}




Figure~\ref{fig:nli-results} shows the performance of the 
MLMs, text-to-text, and causal models of different sizes on ANLI, MNLI, and SNLI;
best results by model class are summarised in Table~\ref{tab:best-results}.
What comes to the fore in this analysis are stark differences between the three datasets, 
visible across all architectures.

\begin{table}[tb!]
\centering
\resizebox{0.48\textwidth}{!}{
\begin{tabular}{@{}cllll@{}}
\toprule
 & \multicolumn{2}{c}{Triplet loss} & \multicolumn{2}{c}{Cosine loss} \\
\multirow{-2}{*}{Task} & \multicolumn{1}{c}{Model} & \multicolumn{1}{c}{Acc} & \multicolumn{1}{c}{Model} & \multicolumn{1}{c}{Acc} \\ \cmidrule(l){2-5} 
 & OPT 30b & 0.701 & Llama 13b & 0.613 \\
 & Llama 3b & 0.675 & OPT 30b & 0.609 \\
\multirow{-3}{*}{ANLI} & ELECTRA l & 0.658 & ELECTRA l & 0.561 \\
\rowcolor[HTML]{EFEFEF} 
\cellcolor[HTML]{EFEFEF} & ELECTRA l & 0.939 & OPT 30b & 0.773 \\
\rowcolor[HTML]{EFEFEF} 
\cellcolor[HTML]{EFEFEF} & RoBERTa l & 0.935 & Llama 7b & 0.764 \\
\rowcolor[HTML]{EFEFEF} 
\multirow{-3}{*}{\cellcolor[HTML]{EFEFEF}SNLI} & OPT 6.7b & 0.929 & RoBERTa l & 0.743 \\
 & RoBERTa l & 0.914 & OPT 30b & 0.86 \\
 & OPT 30b & 0.908 & Llama 13b & 0.856 \\
\multirow{-3}{*}{MNLI} & ELECTRA l & 0.902 & ELECTRA l & 0.821 \\ \bottomrule
\end{tabular}
}
\caption{Best-performing model types (intervening models of the same type but different size were 
skipped) by task and setting. \textit{Acc} stands for accuracy; \textit{l}, for large.}\label{tab:best-results}
\end{table}

\textbf{ANLI} presents the worst results across all model types, albeit with interesting differences.
While the T5-efficient models never do better than random guessing and the MLMs guess randomly in the
cosine-similarity setting and sporadically achieve accuracies of $\approx$\ 0.6 in the triplet setting, 
the models from the Llama and OPT families consistently achieve accuracies above 0.6 in \textit{both}
settings, squarely beating the encoder-equipped models.

The results on \textbf{SNLI}
show the greatest differences between operationalizations.
In the triplet setting, all encoder-based models achieve accuracies of \(\approx\) 0.93 in their lower 
layers, and the results then remain stable or degrade (ELECTRA). 
Causal models attain similar results in the upper layers, and T5-efficient models
demonstrate slightly lower results regardless of the layer. 
In the cosine-similarity setting, however, the task becomes much harder, with no model 
showing accuracy above 0.77, and causal models again showing best performance.

The differences between layers are most pronounced in \textbf{MNLI}. In the triplet setting, MLMs
show the best performance in middle layers, while text-to-text and causal models achieve slightly worse
results in upper layers. In the cosine-similarity setting, however, middle and upper layers of
causal models again demonstrate the best performance, approaching 0.86.

 


On the whole, causal models demonstrate surprisingly good results, outperforming T5-efficient, 
competing with MLMs in the triplet setting, and consistently outperforming them in the more 
challenging cosine-similarity setting. 

\paragraph{Difference between layers} 
The observations on the distribution of information across
model layers made based on the STS task are largely repeated: Llama models tend to achieve peak 
performance in middle layers, while larger OPT models have a dip in performance
between early and late layers, with the best performance attained near the end. The largest OPT
models are also distinguished by an almost uninformative first layer.

\paragraph{Effect of subspace dimensionality} 
Fig.~\ref{fig:nli-results-dim} shows 
that on MNLI and SNLI all models types achieve peak performance with a dimensionality of at least 128
in the triplet-loss setting. This is in contrast to the STS task, where only T5 profited from a dimensionality 
above 64. However, no model can profit from more than 16 dimensions for the cosine-similarity setting,
which highlights
the influence of the finer details of probing methodology on the experimental results.
The results on ANLI are generally inconclusive, as performance is low and unstable throughout;
only OPT-30b seems to systematically gain from dimensionalities above 128.

\paragraph{Effect of model size} 
Up to a certain point, the size of the model is of a much smaller importance than the architecture and 
training regime, and even in the finer details of their performance, differently-sized Llama models 
resemble each other more than the OPT models that are close to them in parameter count. When it comes
to the best performance on the probing task, however, the most successful model is nearly invariably
also the biggest in its class, with the cosine setting being the most size demanding.

\section{Conclusion}

Despite a surge of interest in prompting techniques targeting
large decoder-only language models \citep{liu2023pre}, there are still settings where vector 
representations of sentences remain a competitive alternative, e.g.\ semantic search 
and information retrieval \citep{thakur2021beir,zhuang2023rankt5}. 
Therefore, it seems worthwhile to investigate sentence representations 
from pre-trained models so as to not only better understand models themselves 
but also guide practical applications.


The results of our study suggest two general observations. First, no architecture is best suited for representation 
learning, and the informativeness of vanilla sentence representations can only be measured with regard to a 
particular task. Thus, while the pre-trained RoBERTa
provides the best representations for semantic textual similarity, beating much larger
Llama 13b or OPT 30b and 66b, when it comes to NLI, causal models can provide more informative embeddings
even at smaller model sizes, in line with the findings of \citet{muennighoff2022sgpt} regarding the informativeness of
causal model embeddings for semantic search.

Secondly, different models arrive at very different patterns of information processing across layers. 
Most surprisingly, ELECTRA, despite its similarities to BERT, demonstrates a degradation in performance 
on all surveyed tasks in its upper layers, which begs the question of what kind of linguistic hierarchy 
this model encodes. Similarly, Llama and OPT models, despite sharing the same architecture, also
show markedly differing patterns of information restructuring.

In this study, we targeted two rather general semantics-oriented
tasks.  However, the proposed methodology can be applied to other
problems~-- straightforwardly to regression tasks, such as political
scaling \citep{glavas-etal-2017-unsupervised} or emotion-intensity estimation 
\citep{zad2021emotion}, but also to classification tasks, as long as they support
a reasonable similarity-based reformulation.
We leave the exploration of these areas to future work.


\FloatBarrier

\section*{Limitations}

The results of this study depend on a long series of design choices as to the
particular ways of extracting sentence embeddings, reformulating the downstream
tasks, choosing the loss function, etc. We believe that the choices we made are
justifiable and help to provide a strong lower bound on the informativeness of sentence
representations, but the results we obtained are still dependent on them and different
operationalization may lead to somewhat different conclusions. 

A~more general issue
with this type of analysis is the fact that the notion of \textit{semantics} as encoded
by LMs is not well defined, and while STS and NLI are both reasonable approximations,
there are differences in the way the surveyed models encode information relevant for
these tasks, which, among other things, points to the importance of lexical effects. 
Disentangling these aspects is an important area for future work on model interpretability.

Finally, the validation and test splits of the ANLI dataset in the triplet setting are small,
which leads to noticeable instability of the performance of all models, except for OPT 30b.

\bibliography{anthology,custom}
\bibliographystyle{acl_natbib}

\appendix

\section{Dataset-split sizes}

\subsection{STS task}

Train: 5749; dev: 1500; test: 1379.

\subsection{NLI task}

\subsubsection{Euclidean triplet loss}

\begin{itemize}
    \item \textbf{ANLI} Train: 217940; dev: 116; test: 117.
    \item \textbf{MNLI} Train: 261775; dev: 6692; test: 6703.
    \item \textbf{SNLI} Train: 345241; dev: 3256; test: 3247.
\end{itemize}

\subsubsection{Cosine-similarity loss}

\begin{itemize}
    \item \textbf{ANLI} Train: 94076; dev: 2132; test: 2132.
    \item \textbf{MNLI} Train: 261775; dev: 6692; test: 6703.
    \item \textbf{SNLI} Train: 367384; dev: 6765; test: 6781.
\end{itemize}

\section{Details of model performance}

\begin{figure*}[ht]
    \centering
    \includegraphics[width=0.7\textwidth]{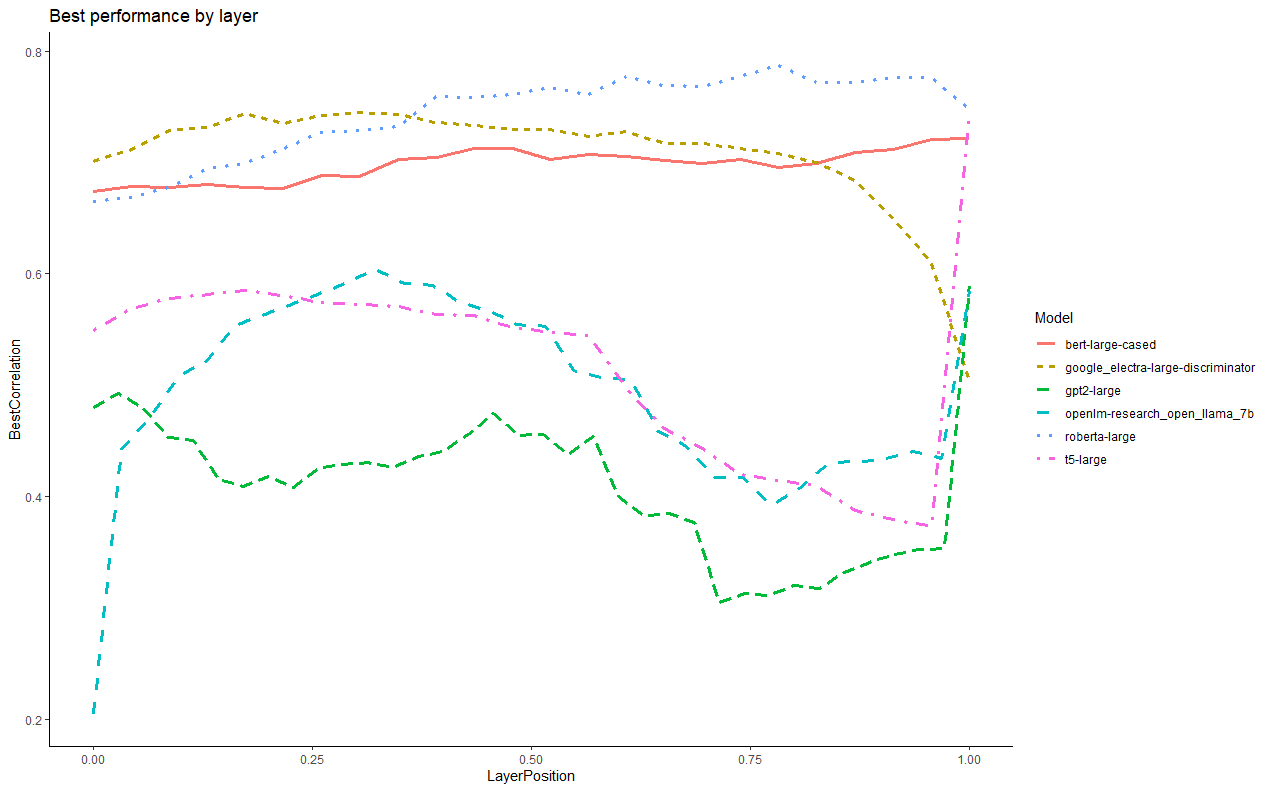}
    \caption{Best performance on the STS task by layer.}
    \label{fig:perf-by-layer}
\end{figure*}

\begin{figure*}[ht]
    \centering
    \includegraphics[width=0.7\textwidth]{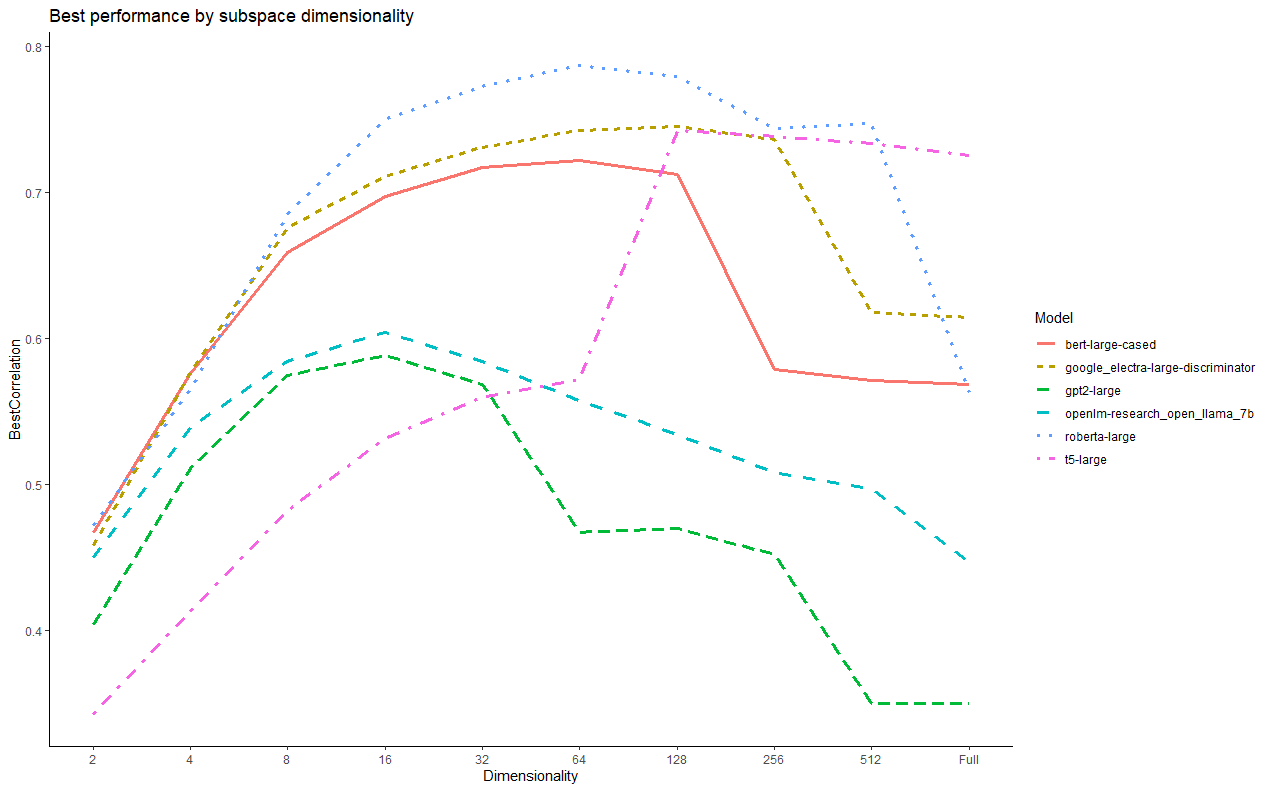}
    \caption{Best performance on the STS task by projection-space dimensionality.}
    \label{fig:perf-by-dimension}
\end{figure*}

Performance of the models on the STS task by layer and by projection-space dimensionality
is shown in Fig.~\ref{fig:perf-by-layer} and \ref{fig:perf-by-dimension} respectively.

\end{document}